\documentclass{article}
\usepackage{iclr2027_conference,times}

\usepackage{amsmath,amsfonts,bm}

\def\eqref#1{equation~\ref{#1}}
\def\1{\bm{1}}

\DeclareMathAlphabet{\mathsfit}{\encodingdefault}{\sfdefault}{m}{sl}
\SetMathAlphabet{\mathsfit}{bold}{\encodingdefault}{\sfdefault}{bx}{n}

\usepackage{enumitem}
\usepackage{graphicx}
\usepackage{url}
\usepackage{tikz}
\usepackage{pgfplots}
\usepackage[section,above,below]{placeins}
\usepackage{booktabs}
\usepackage{xcolor}
\usepackage{array}
\usepackage{caption}
\usepackage{hyperref}
\pgfplotsset{compat=1.18}

\title{RepICL: Reusable In-Context Prediction\\Across Heterogeneous Representation Spaces}

\author{%
  \textbf{Yu-Hsiang Liu}$^{*}$\textbf{,}
  \textbf{Kuan-Yu Chen}$^{\dagger}$\textbf{,}
  \textbf{Chih-Sheng Chen,} \\
  \textbf{Meng-Hsuan Chang,}
  \textbf{Yu-Chen Den \&}  
  \textbf{Tien-Hao Chang} \\
  SinoPac Holdings, Taipei, Taiwan \\
  \small\texttt{\{rickyyhliu,lavamore,sheng77,mandychang,abnerden,darby\}@sinopac.com}
}

\newcommand{\model}[1]{RepICL}
\newcommand{\benchmark}[1]{RepShiftBench}

\iclrfinalcopy

\begin{document}

\maketitle

\renewcommand{\thefootnote}{}
\footnotetext{$^*$Corresponding Author.}
\footnotetext{$^\dagger$Project Leader.}
\renewcommand{\thefootnote}{\arabic{footnote}}

\begin{abstract}
\label{sec:abs}
Frozen representations are widely reused for downstream classification, yet each new task typically requires fitting a new predictor.
We ask whether the few-shot prediction procedure itself can instead be learned once and reused across datasets and representation spaces.
To study this question, we introduce \benchmark{}, comprising 1,218 encoder--dataset tasks across text, image, and audio, with separate evaluation of generalization to unseen datasets, unseen encoders, jointly unseen datasets and encoders, and unseen modalities.
The benchmark exposes a substantial gap: 
Logistic Regression fitted independently on each episode outperforms every evaluated in-context learner across all settings.
We introduce \model{}, a meta-trained in-context learner that canonicalizes each episode through episodic whitening before prediction.
Its inductive variant, \model{}-I, surpasses Logistic Regression in all 12 benchmark settings, while \model{}-T substantially outperforms existing transductive methods.
Ablations identify episodic whitening as the primary source of these gains, while showing that it is not a universally beneficial preprocessing step.
Across both variants, the gains concentrate on queries for which simple support prototypes favor the wrong class or provide little separation between the true class and competing classes.
Transduction provides its largest additional gains when limited support coverage gives a misleading view of class separation.
Together, these results demonstrate that a shared few-shot prediction procedure can generalize beyond the representation spaces observed during training.
\end{abstract}

\section{Introduction}
\label{sec:intro}

Pretrained encoders have made representation reuse a standard paradigm across modern machine learning \citep{Devlin2019BERTPO,radford2021learning,Oquab2024DINOv2LR}.
A frozen encoder can support many downstream datasets, with only a lightweight classifier fitted on top of its representations.
This \emph{embed-then-fit} paradigm is simple and remarkably effective: strong frozen representations often support competitive downstream performance even with basic classifiers \citep{Chen2019CloserLook,kornblith2019better,Tian2020RethinkingFI,ElAssadi2026Embedder}.
Yet this reuse largely stops at the representation:
the downstream prediction rule is still fitted separately for each dataset and refitted when the encoder changes \citep{kornblith2019better,Tian2020RethinkingFI}.

In-context and model-based meta-learning \citep{Santoro2016MANN,Hospedales2021Survey} suggest a more ambitious possibility: rather than repeatedly fitting a task-specific classifier, can we learn the \emph{prediction procedure itself} once and reuse it across tasks?
Recent in-context learners instantiate this idea over vector representations and can transfer across datasets and tasks
\citep{Kirsch2022GeneralPurposeIL,Fifty2023ContextAwareM,Woerner2026UniversalAL}.
However, existing evaluations either keep the encoder fixed or study cross-encoder and cross-modality transfer only in limited settings \citep{Woerner2026UniversalAL}, leaving encoder identity largely unexplored as an independent axis of generalization.
Varying the dataset changes the downstream prediction problem, whereas varying
the encoder changes the representation space itself, as different encoders can induce spaces with different dimensionalities, scales, covariance structures, and geometries.
This raises a distinct question: \emph{\textbf{can a learned prediction procedure still transfer to an unseen representation space?}}

To study this question directly, we introduce \textbf{\benchmark{}}, a benchmark for few-shot classification across heterogeneous frozen representations.
Each task is defined by an \emph{encoder--dataset pair}, allowing the downstream dataset and representation space to vary independently, from which we sample few-shot episodes consisting of support and query examples.
Across 59 datasets and 62 pretrained encoders spanning text, image, and audio, \benchmark{} comprises 1,218 tasks and evaluates generalization to unseen datasets, unseen encoders, jointly unseen datasets and encoders, and entirely unseen modalities. 
This systematic evaluation exposes a substantial gap between reusable prediction procedures and per-episode fitting:
\textbf{\emph{Logistic Regression fitted independently on each episode outperforms every evaluated in-context learner across all benchmark settings}}.
The challenge is therefore not simply few-shot prediction, but learning a shared predictor that remains reliable as the representation space changes.

We address this gap with \textbf{\model{}}, a meta-trained in-context learner that canonicalizes each episode's representations through \emph{episodic whitening}---episode-wise PCA followed by standardization---before in-context prediction.
We study an inductive variant, \model{}-I, which predicts each query from the labeled support set alone, and a transductive variant, \model{}-T, which additionally exploits the unlabeled query set.
\model{}-I surpasses per-episode Logistic Regression in all 12 benchmark settings while consistently outperforming existing in-context learners, and \model{}-T achieves substantially larger gains over existing transductive methods.
Together, these results demonstrate that \emph{\textbf{a single learned prediction procedure can generalize beyond the representation spaces observed during training, including to entirely unseen modalities.}}

Our analyses further clarify what enables this transfer and where its gains arise.
Ablations identify episodic whitening as the primary source of improvement, yet applying the same transformation to existing learners does not improve their performance, indicating that its benefit comes from meta-training the predictor on whitened episodes rather than from better preprocessing alone.
Across both variants, the gains concentrate on queries for which simple support prototypes either favor the wrong class or provide little separation between the true class and competing classes.
Among these difficult queries, transduction provides its largest additional gains when the few-shot support set is an unrepresentative sample of its classes, suggesting that unlabeled queries are especially useful when support coverage is limited.  
\section{Related Work}
\label{sec:related}

\paragraph{Few-shot prediction over pretrained representations.}
Few-shot classification has been studied through metric-based, optimization-based, and model-based approaches~\citep{Vinyals2016MatchingNF,Snell2017PrototypicalNF,Finn2017ModelAgnosticMF,Santoro2016MANN,Hospedales2021Survey}.
With strong pretrained representations, however, simple non-meta-learned rules can be highly competitive: nearest-centroid methods provide strong baselines~\citep{Wang2019SimpleShot}, while linear classifiers fitted on frozen support representations can rival more elaborate meta-learning methods~\citep{Tian2020RethinkingFI}.
Transductive methods additionally exploit unlabeled queries at inference~\citep{Liu2019TPN,Boudiaf2020TIM,Ziko2020LaplacianShot,Hu2021PTMAP}, while
cross-domain few-shot learning studies transfer across data distributions~\citep{Tseng2020CrossDomainFWT,Li2022TSA}.
We instead ask whether a single reusable predictor can retain the strength of per-task fitting as the underlying representation space changes.

\paragraph{In-context prediction across representation spaces.}
Model-based meta-learners can learn reusable prediction procedures over labeled examples and queries without test-time parameter updates ~\citep{Garnelo2018ConditionalNP,Kirsch2022GeneralPurposeIL,Fifty2023ContextAwareM}.
Prior-fitted networks and their tabular extensions similarly learn predictors from distributions of synthetic tasks~\citep{Muller2022PFN,Hollmann2022TabPFNAT,Hollmann2025TabPFNv2,Qu2025TabICLAT}.
Closest to our setting, TAIL targets heterogeneous feature spaces and demonstrates transfer to unseen modalities~\citep{Woerner2026UniversalAL}.
However, its cross-modal evaluation uses only one dataset--encoder pair for each unseen modality, and it does not systematically disentangle dataset and encoder novelty.
Thus, while TAIL establishes the feasibility of transfer across feature spaces and modalities, it remains unclear how a shared predictor behaves across broad combinations of unseen datasets and pretrained encoders. \benchmark{} is designed to study this question systematically.

Methods also differ in how they reduce dependence on feature coordinates.
GPICL meta-trains over randomly projected tasks~\citep{Kirsch2022GeneralPurposeIL}, while TAIL maps encoder outputs through randomized projections into a common latent space~\citep{Woerner2026UniversalAL}.
More broadly, prior work on representation normalization and alignment has normalized episode statistics~\mbox{\citep{Bronskill2020TaskNorm,Hollmann2022TabPFNAT}}, whitened using broader reference statistics~\citep{Su2021Whitening}, or aligned representations through shared anchors~\citep{Moschella2023RelativeRep}.
\model{} instead uses episode-wise PCA and standardization to canonicalize heterogeneous frozen representations for a shared prediction procedure.

\paragraph{Evaluating representation-space generalization.}
Few-shot benchmarks such as Meta-Dataset evaluate learners across diverse classes, datasets, and domains~\citep{Vinyals2016MatchingNF,Triantafillou2020MetaDatasetAD,Ullah2022MetaAlbum}, while embedding benchmarks such as MTEB, MIEB, X-ARES, and MAEB evaluate pretrained representations across broad collections of downstream tasks~\citep{Muennighoff2023MTEBMT,Xiao2025MIEB,Zhang2025XARES,ElAssadi2026MAEB}.
The former primarily vary the downstream task while evaluating few-shot learners, while the latter primarily vary the encoder while evaluating representation quality.
Neither is designed to jointly vary dataset and encoder while evaluating a single shared few-shot predictor.
\benchmark{} fills this gap by treating the encoder--dataset pair as the unit of evaluation and separately evaluating dataset, encoder, joint, and modality generalization.

\section{Few-Shot Prediction Setup}
\label{sec:problem-setup}

\paragraph{Few-shot episodes.}
In this paper, we consider standard $N$-way $K$-shot classification episodes \citep{Vinyals2016MatchingNF,Snell2017PrototypicalNF}.
Each episode contains $N$ classes, with $K$ labeled support examples and $Q$ query examples per class.
Let $\mathcal{S}=\{(\mathbf{z}_i^s,y_i^s)\}_{i=1}^{NK}$ and $\mathcal{Q}=\{(\mathbf{z}_j^q,y_j^q)\}_{j=1}^{NQ}$ denote the support and query sets, respectively, where query labels are withheld from the predictor.
Here, $\mathbf{z}\in\mathbb{R}^d$ denotes an input representation and $y\in\{1,\ldots,N\}$ an episode-local class label, with labels remapped independently within each episode.
The goal is to predict the labels of the query representations from the information available under the corresponding inference setting.

\paragraph{Inductive and transductive inference.}
Let $f_\theta$ denote a few-shot prediction procedure parameterized by $\theta$.
Under inductive inference, each query is predicted independently from the support set and its own representation,
$
\hat{y}_j^q
= f_\theta(\mathcal{S},\mathbf{z}_j^q),
$
where $\hat{y}_j^q$ is the predicted label for query $j$.
Under transductive inference \citep{vapnik1998statistical,Liu2019TPN}, the learner may additionally condition on the full unlabeled query set
$\mathcal{U}=\{\mathbf{z}_j^q\}_{j=1}^{NQ}$ and jointly predict
$
    \hat{\mathbf{y}}^q
    = f_\theta(\mathcal{S},\mathcal{U}),
$
where $\hat{\mathbf{y}}^q$ collects the predictions for all queries.
Because transductive methods access query-set information unavailable to inductive methods, we evaluate the two inference settings separately.
\section{\benchmark{}}
\label{sec:benchmark}
\begin{figure}[t]
\centering
\includegraphics[width=\textwidth]{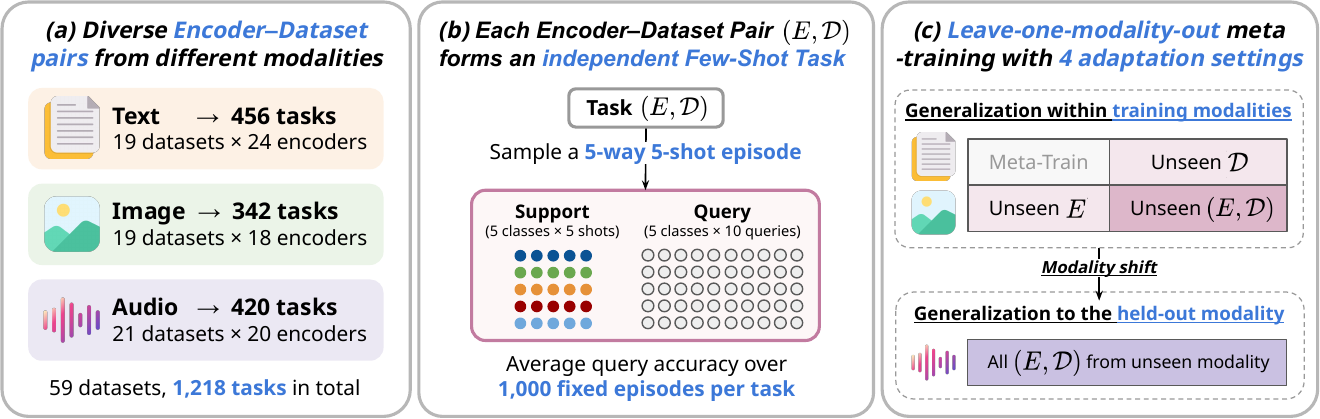}
\caption{\benchmark{}: encoder--dataset tasks, episodic evaluation, and adaptation settings.}
\label{fig:metarepbench-overview}
\end{figure}

The episode formulation in Section~\ref{sec:problem-setup} treats representations as given, without specifying how their representation space varies across tasks.
To study whether a single prediction procedure can generalize when that space itself changes, \benchmark{} treats the pretrained encoder as an explicit task axis alongside the downstream dataset.
Figure~\ref{fig:metarepbench-overview} summarizes the resulting benchmark construction and evaluation protocol.

\subsection{Encoder--Dataset Task Formulation}
\label{subsec:formulation}

\paragraph{Encoder--dataset tasks.}
Let $\mathcal{D}=\{(x_i,y_i)\}_{i=1}^{n}$ denote a labeled dataset and $E$ a frozen pretrained encoder with embedding function $\phi_E:\mathcal{X}\rightarrow\mathbb{R}^{d_E}$.
We define a task as a compatible encoder--dataset pair $(E,\mathcal{D})$.
For each episode sampled from this task, the representations in Section~\ref{sec:problem-setup} are obtained as
$\mathbf{z}=\phi_E(x)\in\mathbb{R}^{d_E}$.
Thus, $\mathcal{D}$ determines the downstream examples and label space, whereas $E$ determines the representation space in which prediction is performed.
Methods receive only these frozen representations and the support labels, rather than the raw inputs.

\paragraph{Generalization settings.}
Let $\mathcal{T}_{\mathrm{train}}$ denote the encoder--dataset tasks used for meta-training.
We evaluate four forms of generalization according to which components of $(E,\mathcal{D})$ are unseen at test time:
\textbf{Dataset Adaptation}, where $\mathcal{D}$ is unseen but $E$ has appeared during meta-training;
\textbf{Encoder Adaptation}, where $E$ is unseen but $\mathcal{D}$ is seen;
\textbf{Joint Adaptation}, where both $E$ and $\mathcal{D}$ are unseen; and
\textbf{Modality Adaptation}, where both are unseen and belong to a modality absent entirely from meta-training.
Together, these settings disentangle changes in the downstream problem from changes in the representation space and extend the evaluation to unseen modalities.

\subsection{Benchmark Construction}
\label{subsec:coverage}

\paragraph{Datasets and encoders.}
\benchmark{} spans text, image, and audio, drawing classification datasets from MTEB for text \citep{Muennighoff2023MTEBMT}, MIEB for images \citep{Xiao2025MIEB}, and X-ARES and MAEB for audio \citep{Zhang2025XARES,ElAssadi2026MAEB}.
We retain datasets with at least five classes and pair each dataset with every compatible encoder of the same modality.
The encoder pool spans diverse architectures, training objectives, model sizes, and embedding dimensions, providing broad variation in the representation spaces encountered by the predictor.
In total, \benchmark{} contains 1,218 compatible encoder--dataset tasks formed from 59 datasets and 62 pretrained encoders.
Complete dataset and encoder inventories and representation extraction details are provided in Appendices~\ref{app:inv-selection} and~\ref{app:inv-repr}.

\paragraph{Folds and task splits.}
We construct three folds, each holding out one modality entirely from meta-training.
Within the remaining two modalities, datasets and encoders are partitioned separately into seen and held-out sets.
Pairs in which both are seen form the meta-training pool, while combinations involving held-out datasets or encoders instantiate the Dataset, Encoder, and Joint Adaptation settings defined above.
The entirely held-out modality defines Modality Adaptation.
Notably, held-out datasets span both unfamiliar domains and familiar task types, while held-out encoders are selected from more recently released models, testing generalization to new datasets and domains and transfer to newer encoders.
Full task assignments and split details are provided in Appendices~\ref{app:split-partitions} and~\ref{app:split-example}.

\subsection{Evaluation Protocol}
\label{subsec:evaluation}

\paragraph{Episode sampling and scoring.} 
Each encoder--dataset task is evaluated on the same 1{,}000 fixed, class-balanced 5-way 5-shot episodes for all methods, with 10 queries per class.
For each evaluation task, supports and queries are drawn according to the source dataset's predefined train--test split whenever available, with supports from the training split and queries from the test split.
Exceptions and split-construction details are provided in Appendix~\ref{app:split-episodes}.
For each task, we average query accuracy across its 1{,}000 episodes.
An adaptation-setting score is computed as the unweighted mean over its encoder--dataset tasks.
When reporting an aggregate score for a modality-held-out fold, we average the four adaptation-setting scores equally, regardless of the number of tasks in each setting.

\paragraph{Reference methods.}
Per-episode methods such as Logistic Regression and ProtoHead are fitted or constructed using only the support set of the current episode and have no meta-training history.
Accordingly, notions such as \emph{seen encoder} or \emph{seen dataset} describe the meta-training history of learned predictors; for per-episode methods, the same adaptation labels simply identify the corresponding evaluation subsets.
All methods are nevertheless evaluated on exactly the same tasks and episodes: per-episode methods quantify what can be achieved from the current support set alone, whereas meta-trained methods test whether knowledge learned across previous tasks and representation spaces provides an additional benefit.
\section{\model{}}
\label{sec:method}
\begin{figure}[t]
\centering
\includegraphics[width=\textwidth]{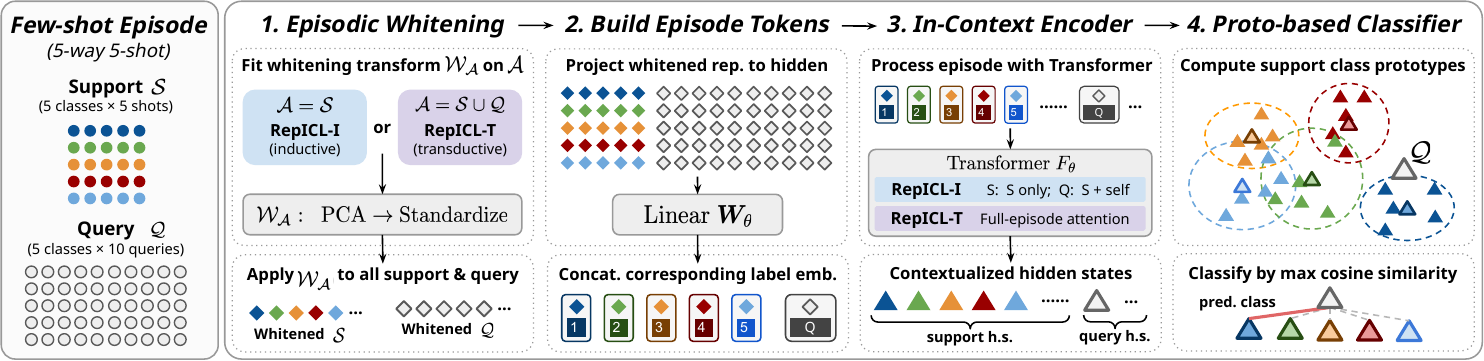}
\caption{\model{}: episodic whitening enables shared in-context prediction across heterogeneous representation spaces.
\model{}-I estimates whitening from the support set and uses inductive attention, where supports attend only to supports and each query attends to supports and itself.
\model{}-T instead uses unlabeled queries for whitening and permits full-episode attention.}
\label{fig:repicl-overview}
\end{figure}

A shared predictor in \benchmark{} must operate across representations produced by different encoders, whose dimensionalities, scales, and covariance structures can vary substantially.
\model{} addresses this heterogeneity by first canonicalizing each episode through \emph{episodic whitening} and then applying a shared Transformer-based in-context predictor in the resulting space, without test-time parameter updates.
We study two inference variants.
\model{}-I is inductive: the labeled support set alone determines both the episode transformation and each query prediction.
\model{}-T is transductive: it additionally uses the unlabeled query set when constructing the episode transformation and during in-context prediction.
Figure~\ref{fig:repicl-overview} summarizes the overall procedure.

\subsection{Episodic Whitening}
\label{subsec:whitening}
Rather than learning an encoder-specific alignment, \model{} constructs an episode-local representation through \emph{episodic whitening}.
For an episode, let $\mathcal{Z}_S=\{\mathbf{z}_i^s\}_{i=1}^{NK}$ and $\mathcal{Z}_Q=\{\mathbf{z}_j^q\}_{j=1}^{NQ}$ denote the support and query representations.
We estimate the episode transformation from $\mathcal{A}=\mathcal{Z}_S$ for \model{}-I and $\mathcal{A}=\mathcal{Z}_S\cup\mathcal{Z}_Q$ for \model{}-T.
We first perform PCA on $\mathcal{A}$ and then standardize the retained
coordinates.
Let $\boldsymbol{\mu}_{\mathcal A}$ be the mean of $\mathcal{A}$ and $\mathbf{W}_{\mathcal A} \in \mathbb R^{d_E\times r_{\mathcal A}}$ the retained principal directions, and $\mathbf m_{\mathcal A}$ and $\mathbf v_{\mathcal A}$ the coordinate-wise mean and variance of the PCA projections of $\mathcal A$.
The episode-specific whitening transform is
\begin{equation}
\mathcal{W}_{\mathcal A}(\mathbf z)
=
\frac{
\mathbf W_{\mathcal A}^{\top}(\mathbf z-\boldsymbol{\mu}_{\mathcal A})-\mathbf m_{\mathcal A}
}{
% \mathbf s_{\mathcal A}+\epsilon
\sqrt{\mathbf v_{\mathcal A}+\varepsilon}
} \in \mathbb{R}^{r_{\mathcal A}},
\end{equation}
where division is elementwise and $\varepsilon>0$ is a small constant for numerical stability.
We set $r_{\mathcal A}=\min(d_E,|\mathcal A|-1,r_{\max})$, with $r_{\max}=24$ for \model{}-I and $74$ for \model{}-T, matching the maximum PCA ranks under the 5-way 5-shot meta-training episode configuration of \benchmark{}.
We then apply $\mathcal{W}_{\mathcal A}$ to all support and query representations, yielding the whitened representations $\mathbf u_i^s$ and $\mathbf u_j^q$ for the shared predictor.

\subsection{In-Context Prediction}
\label{subsec:encoder}

\paragraph{Episode tokens.}
Each whitened representation is projected and concatenated with a label embedding to form a Transformer token:
\begin{equation}
    \mathbf t_i^s =
    [\,\mathbf W_\theta \mathbf u_i^s \,\Vert\, \mathbf e_{y_i^s}\,],
    \qquad
    \mathbf t_j^q =
    [\,\mathbf W_\theta \mathbf u_j^q \,\Vert\, \mathbf e_{q}\,].
\end{equation}
Here, $\mathbf W_\theta$ is a learned projection. Support labels use ELMES embeddings~\citep{Fifty2023ContextAwareM}, while $\mathbf e_q$ is a single learnable embedding shared by all queries.

\paragraph{Contextualization.}
The episode tokens are processed by a shared Transformer $F_\theta$ to obtain contextualized support and query representations $\{\mathbf h_i^s\}$ and $\{\mathbf h_j^q\}$.
For \model{}-I, support tokens attend only to support tokens, while each query attends to all supports and itself but not to other queries.
For \model{}-T, all support and query tokens can attend to one another, allowing information to propagate across the entire episode.

\paragraph{Prototype-based classification.}
The contextualized support representations define an episode-specific classifier.
For each class $c$, let $\mathcal{I}_c=\{i:y_i^s=c\}$ denote its support indices.
We form a class prototype by averaging the corresponding contextualized representations and classify each query by cosine similarity:
\begin{equation}
    \mathbf{p}_c =
    \frac{1}{|\mathcal{I}_c|}
    \sum_{i\in\mathcal{I}_c}\mathbf{h}_i^s,
    \qquad
    \ell_{j,c} =
    \frac{\cos(\mathbf{h}_j^q,\mathbf{p}_c)}{\tau},
    \qquad
    \hat{y}_j^q =
    \arg\max_c \ell_{j,c},
\end{equation}
where $\tau$ is a fixed temperature and $\hat{y}_j^q$ is the predicted label for query $j$.
As in Prototypical Networks~\citep{Snell2017PrototypicalNF} and
FEAT~\citep{Ye2020FEAT}, the classifier is constructed from the current support set rather than tied to fixed class identities.

\subsection{Meta-Training Across Representation Spaces}
\label{subsec:training}
We meta-train a single $\theta$ across tasks on 5-way 5-shot episodes.
We optimize it on episodes sampled from $\mathcal{T}_{\mathrm{train}}$, which contains multiple encoder--dataset pairs, by minimizing query-set cross-entropy:
\begin{equation}
    \mathcal{L}(\theta)
    =
    -\frac{1}{NQ}
    \sum_{j=1}^{NQ}
    \log
    \frac{
        \exp(\ell_{j,y_j^q})
    }{
        \sum_{c=1}^{N}\exp(\ell_{j,c})
    }.
\end{equation}
During meta-training, tasks are sampled to prevent very large datasets from dominating optimization, while class labels are randomly permuted within each episode so that episode-local label embeddings cannot acquire fixed semantic meanings.
By default, we initialize the Transformer and label embeddings from the released CAML checkpoint~\citep{Fifty2023ContextAwareM}, initialize the post-whitening projection from scratch, and replace CAML's fixed output head with the episode-specific prototype classifier described above.
We select the checkpoint with the highest accuracy on the validation tasks. 
Full optimization details are provided in Appendix~\ref{app:model-architecture}.
\section{Experiments}
\label{sec:experiments}

\subsection{Experimental Setup}
\label{subsec:setup}
We evaluate \model{} on \benchmark{} against three groups of baselines:
\textbf{(1) classical classifiers} fitted on each episode's support set, including Logistic Regression and an RBF-kernel SVM;
\textbf{(2) few-shot methods},
including the inductive ProtoHead and SimpleShot~\citep{Wang2019SimpleShot}, and the transductive LaplacianShot~\citep{Ziko2020LaplacianShot}, $\alpha$-TIM~\citep{Veilleux2021RealisticTransductive}, and PT-MAP~\citep{Hu2021PTMAP}; 
and \textbf{(3) in-context learners}, including TAIL~\citep{Woerner2026UniversalAL}, TabPFN~\citep{Hollmann2025TabPFNv2}, and TabICL~\citep{Qu2025TabICLAT}.
Methods trained on \benchmark{} use 5-way 5-shot episodes, and 
all methods are evaluated on the same fixed episodes.
Hyperparameters and training details are provided in Appendix~\ref{app:comparison-details}.

\subsection{Main Results}
\label{subsec:main}

\begin{table*}[t]
\caption{\benchmark{} 5-way 5-shot accuracy (\%) under dataset (D), encoder (E), joint (J), and modality (M) adaptation across three modality-held-out folds. Bold marks the best result within each block. TAIL$^\dagger$ denotes our reproduction of the released recipe with scale calibration (Appendix~\ref{app:tail-reproduction}). Stars indicate a significant improvement over the strongest baseline within the same block and column (two-sided paired test clustered by dataset; $^{*}p<0.05$, $^{**}p<0.01$, $^{***}p<0.001$).}
\centering
\setlength{\tabcolsep}{4pt}
\resizebox{\textwidth}{!}{%
\begin{tabular}{@{}lllllllllllll@{}}
\toprule
\textbf{Adaptation Setting $\rightarrow$} & \multicolumn{4}{c}{\textbf{Text-held-out fold}}
& \multicolumn{4}{c}{\textbf{Audio-held-out fold}}
& \multicolumn{4}{c}{\textbf{Image-held-out fold}} \\
\cmidrule(lr){2-5}\cmidrule(lr){6-9}\cmidrule(l){10-13}
\textbf{Method $\downarrow$} & D & E & J & M & D & E & J & M & D & E & J & M \\
\midrule
\multicolumn{13}{@{}l}{\textbf{\emph{Inductive}}} \\
ProtoHead
& 56.11 & 72.55 & 61.05 & 64.41 & 68.64 & 71.72 & 71.28 & 48.41 & 54.08 & 65.83 & 56.22 & 80.88 \\
SimpleShot
& 56.14 & 73.06 & 61.65 & 64.90 & 68.92 & 72.22 & 72.01 & 49.04 & 54.26 & 66.64 & 57.01 & 81.12 \\
SVM (RBF kernel)
& 55.52 & 72.68 & 62.32 & 64.62 & 67.03 & 72.13 & 71.81 & 50.49 & 55.00 & 66.70 & 57.81 & 76.64 \\
Logistic regression
& 60.19 & 75.33 & 65.74 & 69.62 & 72.37 & 76.05 & 75.56 & 54.46 & 58.35 & 71.56 & 61.90 & 82.38 \\
TabPFN
& 59.26 & 73.32 & 64.36 & 63.38 & 67.96 & 72.46 & 72.71 & 52.55 & 55.30 & 67.31 & 59.19 & 80.65 \\
TabICL
& 59.08 & 74.04 & 63.90 & 67.29 & 70.69 & 74.17 & 73.66 & 52.72 & 56.81 & 69.02 & 59.92 & 81.90 \\
\textbf{\model{}-I (Ours)}
& \textbf{61.41}$^{**}$ & \textbf{75.85}$^{**}$ & \textbf{66.22} & \textbf{70.10}$^{***}$
& \textbf{73.52}$^{**}$ & \textbf{76.79}$^{**}$ & \textbf{75.86} & \textbf{55.14}$^{***}$
& \textbf{59.19}$^{**}$ & \textbf{72.82}$^{***}$ & \textbf{62.56}$^{**}$ & \textbf{83.26}$^{***}$ \\
\midrule
\multicolumn{13}{@{}l}{\textbf{\emph{Transductive}}} \\
LaplacianShot
& 55.21 & 72.78 & 60.55 & 64.42 & 68.77 & 71.39 & 71.22 & 47.53 & 53.48 & 65.71 & 55.61 & 81.15 \\
$\alpha$-TIM
& 59.04 & 74.31 & 64.62 & 66.39 & 70.27 & 73.77 & 73.99 & 52.62 & 56.31 & 68.85 & 59.90 & 82.24 \\
PT-MAP
& 56.82 & 74.60 & 63.00 & 68.65 & 72.04 & 74.47 & 74.82 & 49.46 & 55.73 & 69.92 & 59.29 & 83.01 \\
TAIL$^\dagger$
& 52.79 & 71.96 & 61.05 & 62.23 & 66.06 & 71.13 & 71.07 & 45.67 & 50.33 & 64.93 & 55.55 & 79.31 \\
\textbf{\model{}-T (Ours)}
& \textbf{63.55}$^{**}$ & \textbf{77.99}$^{***}$ & \textbf{68.29}$^{***}$ & \textbf{73.26}$^{***}$
& \textbf{75.27}$^{**}$ & \textbf{78.13}$^{***}$ & \textbf{77.47}$^{**}$ & \textbf{55.62}$^{***}$
& \textbf{61.79}$^{**}$ & \textbf{76.58}$^{***}$ & \textbf{65.02}$^{***}$ & \textbf{85.69}$^{***}$ \\

\bottomrule
\end{tabular}
}
\label{tab:main}
\end{table*}

\paragraph{Results on \benchmark{}.}
Table~\ref{tab:main} reports accuracy across three modality-held-out folds and four adaptation settings.
We report inductive and transductive methods separately because only the latter access the unlabeled query set.
The benchmark exposes a consistent gap between reusable predictors and per-episode fitting:
Logistic Regression is the strongest inductive baseline in all 12 settings, while TabPFN and TabICL remain below it throughout.
This does not imply universal dominance: conventional few-shot methods can outperform Logistic Regression on CNN representations, but this advantage does not persist across broader encoder families or other modalities (Appendix~\ref{app:linear-vs-fewshot}).

Under inductive inference, \model{}-I surpasses Logistic Regression in all 12 settings by
$0.30$--$1.26$ points, with significant gains in 10 of them.
Under transductive inference, \model{}-T leads in all 12 settings, exceeding the strongest transductive baseline by $2.65$--$6.66$ points with significant gains throughout.
Both variants have per-episode runtimes comparable to Logistic Regression (Appendix~\ref{app:inference-timing}).
As an additional comparison, \model{}-I is comparable to LoRA on a subset of \benchmark{} without test-time updates, while \model{}-T further improves with unlabeled queries (Appendix~\ref{app:lora-comparison}).
Together, these results show that a shared prediction procedure can generalize across heterogeneous representation spaces and different shifts under both inference settings.

\paragraph{External validation under the TAIL protocol.}
To test whether these gains depend on the design of \benchmark{}, we additionally evaluate \model{}-T under TAIL's original protocol of 11 visual datasets represented by CLIP features.
We use the image-held-out checkpoint, meta-trained only on text and audio tasks, so all 11 visual datasets and the CLIP image encoder are unseen during meta-training.
Despite this cross-modality transfer, \model{}-T achieves the highest mean accuracy, improving over our TAIL reproduction from 90.47\% to 92.31\%.
Full per-dataset results and comparisons with CAML and GPICL are provided in Appendix~\ref{app:tail-benchmark}, and TAIL reproduction details in Appendix~\ref{app:tail-reproduction}.

\subsection{Ablation Study}
\label{subsec:ablation}

Table~\ref{tab:repicl_ablation} ablates episodic whitening, model-structure choices, and meta-training diversity, with each variant retrained under the same budget across all three folds.
Table~\ref{tab:other_ablation} further tests whether episodic whitening
benefits learners that were not trained on whitened episodes.

\paragraph{Episodic whitening drives the gains.}
Removing either component of episodic whitening causes the largest drops in Table~\ref{tab:repicl_ablation}.
The two components show distinct patterns across adaptation settings: removing coordinate-wise standardization is particularly harmful under dataset adaptation, whereas removing PCA is more damaging when the test encoder is unseen (encoder, joint, and modality adaptation).
Replacing PCA with either a fixed or trainable projection produces similar degradation, suggesting that its benefit comes from the episode-dependent PCA transformation rather than merely matching dimensionality or adding projection capacity.
In contrast, CAML initialization has negligible effect, while replacing the prototype-based classifier with a fixed five-class head is nearly neutral for \model{}-I and slightly improves \model{}-T.
These results identify episodic whitening, rather than initialization or output-head design, as the primary source of \model{}'s gains.

\paragraph{Meta-training diversity primarily benefits transductive inference.}
Restricting meta-training to one dataset or one encoder per modality has only a modest effect on \model{}-I ($0.1$--$0.9$ points), but causes substantially larger drops for \model{}-T ($1.5$--$2.4$ points).
Restricting encoder diversity is also generally more harmful than restricting dataset diversity, suggesting that exposure to varied representation spaces is particularly important for \model{}-T.
Overall, \model{}-I is relatively insensitive to these restrictions, whereas \model{}-T benefits more from diversity across tasks and encoders.

\paragraph{Whitening is not a drop-in preprocessing step.}
To test whether episodic whitening simply produces better inputs, we apply the same transformation to learners not trained on whitened episodes.
Logistic Regression is nearly unchanged ($0.2$--$0.3$ point drops), whereas TabICL loses up to $1.4$ points and TabPFN more than $6$ points.
Thus, whitening does not universally improve representations themselves.
The benefit instead comes from training the shared predictor on whitened episodes.

\begin{table}[t]
\caption{Ablation of \model{} components and meta-training diversity. The full model reports accuracy (\%) averaged across the three folds. All ablations report the change in percentage points relative to the full model, with positive changes bolded. Full fold-level results and the meta-training diversity setup are provided in Appendix~\ref{app:ablation-full}.}
\centering
\resizebox{0.8\textwidth}{!}{%
\begin{tabular}{@{}lrrrrrrrr@{}}
\toprule
& \multicolumn{4}{c}{\textbf{\model{}-I}} & \multicolumn{4}{c}{\textbf{\model{}-T}} \\
\cmidrule(lr){2-5}\cmidrule(l){6-9}
& Dataset & Encoder & Joint & Modality & Dataset & Encoder & Joint & Modality \\
\midrule
\textbf{Full model} & \textbf{64.71} & \textbf{75.15} & \textbf{68.21} & \textbf{69.50} & \textbf{66.87} & \textbf{77.57} & \textbf{70.26} & \textbf{71.52} \\
\midrule
\textbf{\emph{Episodic whitening}} \\
w/o Coord. Standardize & $-$7.06 & $-$4.77 & $-$4.81 & $-$6.84 & $-$11.41 & $-$9.37 & $-$9.78 & $-$10.24 \\
w/o PCA, trainable proj. & $-$6.38 & $-$9.90 & $-$9.42 & $-$9.52 & $-$7.55 & $-$11.61 & $-$10.80 & $-$11.22 \\
w/o PCA, fixed proj. & $-$6.37 & $-$9.51 & $-$9.46 & $-$9.13 & $-$8.02 & $-$11.98 & $-$11.03 & $-$11.24 \\
\midrule
\textbf{\emph{Model structure}} \\
w/o CAML initialization & $-$0.07 & $-$0.12 & \textbf{+0.11} & $-$0.02 & \textbf{+0.34} & \textbf{+0.35} & \textbf{+0.21} & \textbf{+0.47} \\
fixed five-class head & $-$0.21 & $-$0.32 & \textbf{+0.13} & $-$0.08 & \textbf{+1.01} & \textbf{+0.91} & \textbf{+0.90} & \textbf{+1.18} \\
\midrule
\textbf{\emph{Meta-training diversity}} \\
one dataset per modality & $-$0.44 & $-$0.47 & $-$0.12 & $-$0.42 & $-$1.91 & $-$1.97 & $-$1.76 & $-$1.69 \\
one encoder per modality & $-$0.87 & $-$0.70 & $-$0.30 & $-$0.49 & $-$2.40 & $-$2.35 & $-$2.00 & $-$1.52 \\
\bottomrule
\end{tabular}
}
\label{tab:repicl_ablation}
\end{table}

\begin{table}[t]
\caption{Accuracy (\%) on raw representations and the change after applying episodic whitening, averaged across the four adaptation settings within each modality-held-out fold. Results use 30 episodes per task due to compute constraints, so raw accuracies differ slightly from Table~\ref{tab:main}.}
\centering
\resizebox{0.8\textwidth}{!}{%
\begin{tabular}{@{}lccccccccc@{}}
\toprule
& \multicolumn{3}{c}{\textbf{Logistic Regression}} & \multicolumn{3}{c}{\textbf{TabICL}}& \multicolumn{3}{c}{\textbf{TabPFN}} \\
\cmidrule(lr){2-4}\cmidrule(l){5-7}\cmidrule(l){8-10}
& Text & Audio & Image & Text & Audio & Image & Text & Audio & Image \\
\midrule
Raw representations & 67.44 & 69.59 & 68.36 & 65.70 & 67.81 & 66.59 & 64.66 & 66.42 & 65.37 \\
+ Episodic whitening & -0.32 & -0.34 & -0.22 & $-$1.05 & $-$1.42 & $-$1.11 & $-$5.91 & $-$6.27 & $-$6.22 \\
\bottomrule
\end{tabular}
}
\label{tab:other_ablation}
\end{table}

\subsection{Analyses}
\label{subsec:analyses}

We use support-prototype geometry to answer two questions raised by the main results: where does inductive \model{} improve over per-episode Logistic Regression, and among these difficult cases, when does transductive inference provide additional benefit?

\paragraph{Where does \model{}-I improve over Logistic Regression?}
For each query, we form class prototypes from the 5-shot support set in the unit-normalized representation space.
Let $D_{\mathrm{own}}$ denote the squared distance to the prototype of the ground-truth class and $D_{\mathrm{other}}$ the distance to the nearest competing prototype.
We define the prototype margin
\begin{equation}
m =
\frac{D_{\mathrm{other}}-D_{\mathrm{own}}}
     {D_{\mathrm{other}}+D_{\mathrm{own}}}.
\label{eq:margin}
\end{equation}
A positive margin indicates that the ground-truth prototype is nearest, whereas $m<0$ indicates that a competing prototype is closer; we group $m\le0$ as queries not correctly resolved by the support prototypes.

Table~\ref{tab:query-margin} reveals a sharp difference between these regimes.
For queries with $m\le0$, \model{}-I improves over Logistic Regression by 2.89 points and outperforms it across all negative-margin bins.
In contrast, Logistic Regression is slightly stronger when $m>0$, with its advantage becoming more consistent as the margin increases.
The same pattern is amplified by \model{}-T.
Thus, across both variants, the gains concentrate on queries for which simple support prototypes either favor the wrong class or provide little separation between the true class and competing classes.

\begin{table}[t]
\caption{Accuracy (\%) by prototype margin $m$, pooled across the three modality-held-out folds (query-weighted).
The first two rows aggregate queries by the sign of $m$, while the remaining rows provide finer margin bins.
Bold marks the best method in each row.}
\label{tab:query-margin}
\centering
\small
\resizebox{0.8\textwidth}{!}{%
\begin{tabular}{@{}lccccccc@{}}
\toprule
Margin $m$ & Queries (\%) & Logistic regression & TabICL & TAIL & PT-MAP & \model{}-I & \model{}-T \\
\midrule
$m\le 0$          & 36.3 & 24.66 & 22.19 & 18.22 & 22.34 & 27.55 & \textbf{34.27} \\
$m>0$             & 63.7 & \textbf{92.78} & 91.59 & 86.39 & 91.15 & 92.33 & 91.86 \\
\midrule
$<-0.20$          & 10.5 & 17.84 & 13.54 & 7.46 & 9.65 & 22.95 & \textbf{25.61} \\
$[-0.20,-0.10)$   &  8.2 & 19.92 & 16.74 & 12.25 & 15.45 & 23.50 & \textbf{28.72} \\
$[-0.10,-0.05)$   &  7.2 & 23.27 & 21.32 & 18.78 & 22.75 & 25.54 & \textbf{33.47} \\
$[-0.05,-0.02)$   &  5.9 & 29.80 & 29.82 & 28.37 & 33.50 & 30.51 & \textbf{42.08} \\
$[-0.02,0]$       &  4.6 & 44.42 & 43.40 & 39.71 & 48.81 & 44.73 & \textbf{55.27} \\
\addlinespace
$(0,0.02)$        &  4.8 & 66.56 & 59.70 & 51.71 & 64.99 & 68.59 & \textbf{69.10} \\
$[0.02,0.05)$     &  6.9 & 83.35 & 77.24 & 66.28 & 78.72 & \textbf{83.40} & 81.12 \\
$[0.05,0.10)$     & 10.0 & \textbf{91.26} & 89.93 & 81.26 & 87.87 & 90.24 & 89.18 \\
$[0.10,0.20)$     & 14.1 & 95.57 & \textbf{95.88} & 90.69 & 93.93 & 94.50 & 94.14 \\
$\ge 0.20$        & 27.9 & 98.67 & \textbf{98.95} & 96.88 & 98.42 & 98.20 & 98.15 \\
\bottomrule
\end{tabular}
}
\end{table}

\paragraph{What does transductive inference add?}
A non-positive margin can arise for different reasons: the 5-shot support set may provide an unrepresentative sample of its classes, or the query may remain ambiguous even when more labeled examples are available.
To distinguish these cases, we recompute Equation~\ref{eq:margin} using leave-one-query-out prototypes formed from all other labeled examples in the episode, including the remaining queries, and denote the resulting \emph{oracle prototype margin} by $m^{\star}$.
Because $m^{\star}$ uses query labels, it is unavailable to all evaluated methods and serves only as an analysis diagnostic.
Relative to the 5-shot margin $m$, it provides a less support-sensitive reference for the episode geometry.

Table~\ref{tab:support-draw} shows that the benefit of transduction depends strongly on the relationship between the 5-shot and oracle margins.
When $m\le0$ but $m^{\star}>0$, the 5-shot support prototypes favor a competing class, whereas prototypes estimated from the larger labeled reference set favor the ground-truth class.
In this regime, \model{}-T improves over \model{}-I by 12.56 points.
The gain is much smaller when both margins remain non-positive, becomes negligible when both are positive, and reverses when the support prototypes favor the correct class but the oracle margin does not.
This asymmetry suggests that unlabeled query context is especially useful when limited support coverage gives a misleading view of class separation, rather than uniformly improving prediction across queries.

\begin{table}[t]
\caption{Accuracy (\%) stratified by the signs of the 5-shot prototype margin $m$ and oracle prototype margin $m^{\star}$, pooled across the three modality-held-out folds (query-weighted).
Bold marks the best method in each row.
T $-$ I reports the accuracy difference between \model{}-T and \model{}-I.}
\label{tab:support-draw}
\centering
\small
\setlength{\tabcolsep}{3.5pt}
\resizebox{0.8\textwidth}{!}{%
\begin{tabular}{@{}cccccccccr@{}}
\toprule
$m$ & $m^{\star}$ & Queries (\%) & Logistic regression & TabICL & TAIL & PT-MAP & \model{}-I & \model{}-T & T $-$ I \\
\midrule
$\le 0$ & $\le 0$ & 22.9 & 21.24 & 18.61 & 14.53 & 14.79 & 23.89 & \textbf{27.18} & $\mathbf{+3.29}$\\
$\le 0$ & $> 0$ & 13.5 & 30.48 & 28.29 & 24.50 & 35.17 & 33.75 & \textbf{46.32} & $\mathbf{+12.56}$\\
$> 0$ & $\le 0$ & 10.0 & \textbf{78.47} & 75.81 & 66.17 & 68.38 & 76.19 & 72.13 & $\mathbf{-4.06}$\\
$> 0$ & $> 0$ & 53.7 & 95.44 & 94.53 & 90.15 & 95.39 & 95.34 & \textbf{95.53} & $+0.19$\\
\bottomrule
\end{tabular}
}
\end{table}
\section{Conclusion, Limitations and Future Work}
\label{sec:conclusion}

\paragraph{Conclusion.} 
In this paper, we studied whether a few-shot prediction procedure, like pretrained representations, can be learned once and reused across heterogeneous frozen representation spaces.
\benchmark{} exposes a substantial gap in this setting: existing in-context learners consistently trail simple per-episode Logistic Regression.
\model{} shows that a shared predictor can nevertheless surpass the strongest per-episode baseline across changing datasets and representation spaces.
Its inductive variant surpasses Logistic Regression across all benchmark settings, while \model{}-T achieves larger gains under transductive inference.
Our analyses identify episodic whitening as the primary source of these improvements and show that \model{}'s gains concentrate on queries that are poorly resolved by simple support prototypes, with transduction providing its largest additional benefit when support coverage is limited.

\paragraph{Limitations and future work.} Our evaluation focuses on 5-way 5-shot classification, leaving generalization to other episode shapes, including different numbers of ways or shots, for future work.
Moreover, while \benchmark{} varies representation spaces across tasks, each episode is still drawn from a single encoder--dataset pair.
These limitations motivate two natural extensions: improving generalization across episode shapes, and moving from transfer \emph{across} representation spaces to joint reasoning \emph{over multiple} representation spaces within an episode.
These directions point toward a broader goal: a reusable prediction procedure that could serve as a common adaptation layer between pretrained encoders and downstream tasks.
Rather than fitting a new predictor whenever the dataset, encoder, or modality changes, such a procedure could amortize adaptation across them while exploiting richer episode-level structure than simple per-task fitting.

\section*{AI use statement}
We used generative AI tools to discuss research ideas and experiments, write code, draft and polish the paper, and find related work, and did not use them for any other task that requires disclosure under the ICLR 2027 AI policy.
We have reviewed all AI-assisted work: code, results, and claims were checked against the recorded experiments, and literature suggestions were verified against the cited sources. We take responsibility for the final content of this work, including text, claims, code, and analyses produced with the aid of generative AI.

\section*{Reproducibility statement}
To support reproducibility, we describe our setup in the main paper and give further details in the appendices.
Section~\ref{sec:problem-setup} introduces the few-shot setup, and Section~\ref{sec:benchmark} introduces \benchmark{}, whose construction is detailed in Appendix~\ref{app:task-inventory}. 
Section~\ref{sec:method} describes \model{}, and Appendix~\ref{app:model-architecture} gives its training details. The compared methods are introduced in Section~\ref{subsec:setup} and detailed in Appendix~\ref{app:comparison-details}, and our reproduction of TAIL is described in Appendix~\ref{app:tail-reproduction}.

\bibliography{iclr2027_conference}
\bibliographystyle{iclr2027_conference}

\appendix
\section{\benchmark{} Construction}
\label{app:task-inventory}
\subsection{Datasets and Encoders}
\label{app:inv-selection}
\paragraph{Datasets.}
\benchmark{} covers three modalities: text, image, and audio. For each modality, we draw datasets from established embedding benchmarks that span diverse application domains and classification tasks: MTEB for text, MIEB for images, and X-ARES and MAEB for audio. We select single-label, example-level classification tasks that support our five-way episodic protocol; this excludes, for example, multi-label and frame-level annotation tasks. In addition, we adapt six MTEB clustering datasets for classification by treating their cluster labels as classes.

\paragraph{Encoders.}
We select encoders to capture diversity across model families, training objectives, and release periods. The inventory includes models commonly used for classification as well as models designed to produce general-purpose embeddings. Their public release dates range from 2014 to 2026. When a model family has several variants, we include one variant to limit redundancy. Tables~\ref{tab:MetaRepBench-datasets} and~\ref{tab:encoder-inventory} list the full inventory.

\subsection{Representation Extraction}
\label{app:inv-repr}
\label{app:inv-preproc}
\paragraph{Preprocessing.}
Every encoder receives inputs preprocessed as it requires, such as image resizing and normalization or audio resampling. Beyond this, we apply a few dataset-specific steps. For text, the MTEB clustering datasets store each record as a list of texts with their cluster labels, so we expand each list into one example per text and remove exact duplicates. We also exclude text examples longer than 2{,}048 tokens, measured with a single fixed tokenizer (GPT-2), so that every text encoder sees the same examples. Due to limited GPU resources, we further truncate the inputs of decoder-only LLMs used through their hidden states, GritLM, and LLM2Vec to 512 tokens, and keep only the first 30 seconds of each VoxCeleb1 and BirdCLEF clip.

\paragraph{Embedding extraction.}
Each frozen encoder then maps the preprocessed inputs to one embedding per example, which we compute once and cache. The embedding is obtained by selecting a designated token or averaging the encoder's outputs over tokens, patches, or frames, depending on the modality, and is then L2-normalized. We generally follow each encoder's released embedding pipeline where one exists.

\subsection{Folds, Splits, and Adaptation Settings}
\label{app:split-partitions}
\paragraph{Folds.}
The three folds of Section~\ref{subsec:coverage} let us compare meta-trained methods fairly against few-shot methods that receive no meta-training. Each fold holds out one of the three modalities (text, image, or audio) and draws its meta-training tasks only from the other two. Together with the dataset and encoder splits below, this keeps every test task out of meta-training, so a meta-trained method is never tested on a task it was trained on. Rotating the held-out modality across the three folds tests generalization on every modality, not just one.

\paragraph{Dataset and encoder splits.}
Within each modality, datasets and encoders are separately partitioned into seen and held-out sets, and this partition is shared by all folds. In the fold where a modality is held out, none of its datasets or encoders is used for meta-training, including those in its seen set. For encoders, the held-out set contains the most recently released encoders of the modality, so that Encoder Adaptation tests transfer to newer models. For datasets, the held-out set includes both out-of-domain datasets, such as remote-sensing images and speaker counting, and datasets whose task type also appears in the seen set. This tests generalization to unfamiliar domains as well as to new datasets of familiar task types. Tables~\ref{tab:MetaRepBench-datasets} and~\ref{tab:encoder-inventory} give the split of every dataset and encoder.

\paragraph{Task sets and adaptation settings.}
In each fold, every task belongs to the meta-training, validation, or test set. In the two meta-training modalities, the tasks whose dataset and encoder are both seen, which form the meta-training pool of Section~\ref{subsec:coverage}, are divided into a meta-training set $\mathcal{T}_{\mathrm{train}}$ (Section~\ref{subsec:formulation}) and a validation set, with about 20\% of them in the latter for model and hyperparameter selection. All other tasks form the test set, and each belongs to one of the four adaptation settings of Section~\ref{subsec:formulation}. A task of the held-out modality is a Modality Adaptation task. A task of a meta-training modality is a Dataset, Encoder, or Joint Adaptation task, depending on whether its dataset, its encoder, or both are held out. The four adaptation settings represent different levels of generalization: an unseen dataset, an unseen encoder, both, or an unseen modality.

The same task can fall into different adaptation settings in different folds. A text task with a seen dataset and a held-out encoder, for example, is an Encoder Adaptation task when image is held out, but a Modality Adaptation task when text is held out. The adaptation settings describe what a meta-trained method has seen before testing, so in Table~\ref{tab:main} they apply only to \model{}. For the other methods, the folds and adaptation settings only determine which tasks they are evaluated on and how these tasks are grouped. Evaluating \model{} in each adaptation setting shows whether its ability is limited to particular adaptation settings or is consistent across all of them.

\subsection{Episode Construction}
\label{app:split-example}
\label{app:split-episodes}

\paragraph{Episode sampling.}
\benchmark{} defines all evaluation episodes as 5-way 5-shot with 10 queries per class, and we use the same shape for meta-training and validation. Each episode is sampled from a single task: we sample 5 classes, then 5 support and 10 query examples per class, all without replacement, and randomly permute the class labels in each episode. Episodes therefore use only classes with enough examples for this shape, and a task needs at least five such classes. Table~\ref{tab:MetaRepBench-datasets} reports, for every dataset, the number of classes that meet this requirement and the number of all classes.

\paragraph{Example pools.}
Each dataset has a support pool and a query pool of examples, shared by all encoders. Both follow the splits of the source benchmark: the support pool is its original training split and the query pool its test split. If the source lacks such a split, we create it by class-stratified splitting; for a few sources that provide a single labeled split, both pools are that split and support and queries are sampled without overlap. Episodes of tasks in the test set sample 5 support examples per class from the support pool and 10 queries per class from the query pool. Meta-training and validation episodes sample all 15 examples per class from the support pool, so the query pool is never used for meta-training or validation.

\paragraph{Fixed evaluation episodes.}
We sample the 1{,}000 evaluation episodes of each task in the test set once and use them, including their label assignment, for all methods. A task also keeps the same episodes when it appears in different folds and adaptation settings, so differences across settings do not come from different episodes.

\begin{table}[p]
\caption{Datasets of \benchmark{}, grouped by modality and by seen/held-out set (Appendix~\ref{app:split-partitions}). \#Classes gives available and all classes (Appendix~\ref{app:split-episodes}).}
\label{tab:MetaRepBench-datasets}
\centering
\setlength{\tabcolsep}{4pt}
\resizebox{0.98\textwidth}{!}{%
\begin{tabular}{@{}llrrl@{}}
\toprule
Dataset & Category & \#Classes (avail./all) & \#Examples & Source \\
\midrule
\multicolumn{5}{@{}l}{\textit{Text} (seen)}\\
ArXivHierarchicalClusteringS2S & Academic-paper clustering & 169 / 169 & 627,949 & MTEB\\
Banking77Classification & Intent classification & 77 / 77 & 13,069 & MTEB\\
EmotionClassification & Emotion classification & 6 / 6 & 19,930 & MTEB\\
HWU64 & Intent classification & 57 / 64 & 10,030 & MTEB\\
MassiveIntentClassification & Intent classification & 54 / 60 & 16,521 & MTEB\\
MassiveScenarioClassification & Scenario classification & 18 / 18 & 16,521 & MTEB\\
MTOPDomainClassification & Domain classification & 11 / 11 & 19{,}723 & MTEB\\
MTOPIntentClassification & Intent classification & 32 / 113 & 19{,}723 & MTEB\\
RedditClustering & Topic clustering & 50 / 50 & 64,337 & MTEB\\
StackExchangeClustering & Topic clustering & 171 / 171 & 514,513 & MTEB\\
TREC (coarse) & Question-type classification & 5 / 6 & 5,952 & MTEB\\
TweetTopicSingleClassification & Topic classification & 6 / 6 & 3,209 & MTEB\\
TwentyNewsgroupsClustering & Topic clustering & 20 / 20 & 5,242 & MTEB\\
YahooAnswersTopicsClassification & Topic classification & 10 / 10 & 1,341,064 & MTEB\\
\multicolumn{5}{@{}l}{\hspace{1em}\textit{held-out}}\\
AmazonReviewsClassification & Review sentiment classification & 5 / 5 & 210,000 & MTEB\\
BiorxivClusteringS2S & Academic-paper clustering & 25 / 26 & 53,773 & MTEB\\
CLINC150 (no OOS) & Intent classification & 150 / 150 & 22,500 & MTEB\\
DBpediaClassification & Topic classification & 14 / 14 & 4,096 & MTEB\\
MedrxivClusteringS2S & Academic-paper clustering & 49 / 51 & 17,646 & MTEB\\
\midrule
\multicolumn{5}{@{}l}{\textit{Image} (seen)}\\
BirdsnapClassification & Fine-grained bird classification & 500 / 500 & 17,849 & MIEB\\
Caltech101Classification & Generic object classification & 87 / 102 & 9,144 & MIEB\\
CIFAR10Classification & Generic object classification & 10 / 10 & 60,000 & MIEB\\
CIFAR100Classification & Generic object classification & 100 / 100 & 60,000 & MIEB\\
Country211Classification & Geolocation classification & 211 / 211 & 52,750 & MIEB\\
DTDClassification & Texture classification & 47 / 47 & 5,640 & MIEB\\
FER2013Classification & Facial-expression classification & 7 / 7 & 35,887 & MIEB\\
Food101Classification & Fine-grained food classification & 101 / 101 & 101,000 & MIEB\\
ImageNet1KClassification & Generic object classification & 1000 / 1000 & 1,331,167 & MIEB\\
MNISTClassification & Handwritten-digit classification & 10 / 10 & 70,000 & MIEB\\
OxfordFlowersClassification & Fine-grained flower classification & 83 / 102 & 8,189 & MIEB\\
OxfordPetsClassification & Fine-grained pet classification & 37 / 37 & 7,349 & MIEB\\
StanfordCarsClassification & Fine-grained car classification & 196 / 196 & 16,185 & MIEB\\
STL10Classification & Generic object classification & 10 / 10 & 13,000 & MIEB\\
SUN397Classification & Scene classification & 397 / 397 & 108,752 & MIEB\\
\multicolumn{5}{@{}l}{\hspace{1em}\textit{held-out}}\\
EuroSATClassification & Remote-sensing classification & 10 / 10 & 27,000 & MIEB\\
FGVCAircraftClassification & Fine-grained aircraft classification & 100 / 100 & 10,000 & MIEB\\
GTSRBClassification & Traffic-sign classification & 43 / 43 & 39,270 & MIEB\\
RESISC45Classification & Remote-sensing classification & 45 / 45 & 31,500 & MIEB\\
\midrule
\multicolumn{5}{@{}l}{\textit{Audio} (seen)}\\
CREMA-D & Speech-emotion classification & 6 / 6 & 7,442 & X-ARES\\
ESC-50 & Environmental-sound classification & 50 / 50 & 2,000 & X-ARES\\
Fluent Speech Commands & Spoken-intent classification & 247 / 248 & 30,043 & X-ARES\\
Free Music Archive Small & Music-genre classification & 8 / 8 & 7,994 & X-ARES\\
FSD18-Kaggle & Sound-event classification & 41 / 41 & 11,073 & X-ARES\\
GTZAN Genre & Music-genre classification & 9 / 10 & 999 & X-ARES\\
MInDS-14 & Spoken-intent classification & 14 / 14 & 8,168 & MAEB\\
NSynth-Instruments & Musical-instrument classification & 10 / 11 & 71,774 & X-ARES\\
SIB-FLEURS & Spoken-topic classification & 7 / 7 & 11,424 & MAEB\\
Speech Commands V1 & Keyword spotting & 30 / 30 & 64,721 & X-ARES\\
VocalSound & Vocal-sound classification & 6 / 6 & 20,985 & X-ARES\\
VoxCeleb1 & Speaker identification & 94 / 1251 & 153,412 & X-ARES\\
VoxLingua33 & Language identification & 24 / 33 & 67,609 & X-ARES\\
VoxPopuliLanguageID & Language identification & 5 / 5 & 500 & MAEB\\
\multicolumn{5}{@{}l}{\hspace{1em}\textit{held-out}}\\
BirdCLEF & Bioacoustic classification & 50 / 50 & 1,000 & MAEB\\
CommonLanguageAgeDetection & Speaker-age classification & 5 / 8 & 6,000 & MAEB\\
LibriCount & Speaker-count classification & 11 / 11 & 5,720 & X-ARES\\
MridinghamTonic & Music-tonic classification & 6 / 6 & 6,977 & MAEB\\
RAVDESS & Speech-emotion classification & 8 / 8 & 1,440 & X-ARES\\
UrbanSound 8K & Environmental-sound classification & 10 / 10 & 8,732 & X-ARES\\
Vocal Imitation & Vocal-imitation classification & 9 / 302 & 5,601 & X-ARES\\
\bottomrule
\end{tabular}
}
\end{table}

\begin{table}[p]
\caption{Encoders of \benchmark{}, grouped by modality and by seen/held-out set (Appendix~\ref{app:split-partitions}), and ordered within each group by the release date of the cited paper. Checkpoints are from Hugging Face unless marked $^*$, which come from other sources. $^\dagger$SigLIP~2 (2025-02) takes the release date and split of the original SigLIP.}
\label{tab:encoder-inventory}
\centering
\scriptsize
\setlength{\tabcolsep}{2pt}
\renewcommand{\arraystretch}{0.88}
\resizebox{\textwidth}{!}{%
\begin{tabular}{@{}lcrl@{}}
\toprule
Checkpoint & Release & Dim. & Ref. \\
\multicolumn{4}{@{}l}{\textit{Text} (seen)}\\
\href{https://nlp.stanford.edu/projects/glove/}{\texttt{stanfordnlp/glove.6B.300d}}$^*$ & 2014-10 & 300 & \citet{Pennington2014GloVe} \\
\href{https://huggingface.co/facebook/fasttext-en-vectors}{\texttt{facebook/fasttext-en-vectors}} & 2018-02 & 300 & \citet{Grave2018FastText} \\
\href{https://huggingface.co/google-bert/bert-base-uncased}{\texttt{google-bert/bert-base-uncased}} & 2018-10 & 768 & \citet{Devlin2019BERTPO} \\
\href{https://huggingface.co/openai-community/gpt2}{\texttt{openai-community/gpt2}} & 2019-02 & 768 & \citet{Radford2019GPT2} \\
\href{https://huggingface.co/sentence-transformers/all-mpnet-base-v2}{\texttt{sentence-transformers/all-mpnet-base-v2}} & 2019-08 & 768 & \citet{reimers2019sentence} \\
\href{https://huggingface.co/openai/clip-vit-large-patch14}{\texttt{openai/clip-vit-large-patch14}} & 2021-02 & 768 & \citet{radford2021learning} \\
\href{https://huggingface.co/microsoft/deberta-v3-large}{\texttt{microsoft/deberta-v3-large}} & 2021-11 & 1024 & \citet{He2021DeBERTaV3} \\
\href{https://huggingface.co/intfloat/e5-large-v2}{\texttt{intfloat/e5-large-v2}} & 2022-12 & 1024 & \citet{Wang2022E5} \\
\href{https://huggingface.co/google/siglip-so400m-patch14-224}{\texttt{google/siglip-so400m-patch14-224}} & 2023-03 & 1152 & \citet{Zhai2023SigLIP} \\
\href{https://huggingface.co/BAAI/bge-large-en-v1.5}{\texttt{BAAI/bge-large-en-v1.5}} & 2023-09 & 1024 & \citet{Xiao2023CPack} \\
\href{https://huggingface.co/mistralai/Mistral-7B-Instruct-v0.3}{\texttt{mistralai/Mistral-7B-Instruct-v0.3}} & 2023-10 & 4096 & \citet{Jiang2023Mistral} \\
\href{https://huggingface.co/intfloat/e5-mistral-7b-instruct}{\texttt{intfloat/e5-mistral-7b-instruct}} & 2023-12 & 4096 & \citet{Wang2024E5Mistral} \\
\href{https://huggingface.co/GritLM/GritLM-7B}{\texttt{GritLM/GritLM-7B}} & 2024-02 & 4096 & \citet{Muennighoff2024GritLM} \\
\href{https://huggingface.co/McGill-NLP/LLM2Vec-Mistral-7B-Instruct-v2-mntp-supervised}{\texttt{McGill-NLP/LLM2Vec-Mistral-7B-Instruct-v2-mntp-supervised}} & 2024-04 & 4096 & \citet{BehnamGhader2024LLM2Vec} \\
\href{https://huggingface.co/nvidia/NV-Embed-v2}{\texttt{nvidia/NV-Embed-v2}} & 2024-05 & 4096 & \citet{Lee2024NVEmbed} \\
\href{https://huggingface.co/meta-llama/Meta-Llama-3.1-8B-Instruct}{\texttt{meta-llama/Meta-Llama-3.1-8B-Instruct}} & 2024-07 & 4096 & \citet{Grattafiori2024Llama3} \\
\href{https://huggingface.co/BAAI/bge-en-icl}{\texttt{BAAI/bge-en-icl}} & 2024-09 & 4096 & \citet{Li2024BGEICL} \\
\href{https://huggingface.co/NovaSearch/stella_en_1.5B_v5}{\texttt{NovaSearch/stella\_en\_1.5B\_v5}} & 2024-12 & 1024 & \citet{Zhang2024Stella} \\
\href{https://huggingface.co/answerdotai/ModernBERT-base}{\texttt{answerdotai/ModernBERT-base}} & 2024-12 & 768 & \citet{Warner2024ModernBERT} \\
\multicolumn{4}{@{}l}{\hspace{1em}\textit{held-out}}\\
\href{https://huggingface.co/deepseek-ai/DeepSeek-R1-Distill-Qwen-7B}{\texttt{deepseek-ai/DeepSeek-R1-Distill-Qwen-7B}} & 2025-01 & 3584 & \citet{DeepSeek2025R1} \\
\href{https://huggingface.co/google/gemma-3-4b-it}{\texttt{google/gemma-3-4b-it}} & 2025-03 & 2560 & \citet{Gemma2025Gemma3} \\
\href{https://huggingface.co/Qwen/Qwen3-8B}{\texttt{Qwen/Qwen3-8B}} & 2025-05 & 4096 & \citet{Yang2025Qwen3} \\
\href{https://huggingface.co/Qwen/Qwen3-Embedding-8B}{\texttt{Qwen/Qwen3-Embedding-8B}} & 2025-06 & 4096 & \citet{Zhang2025Qwen3Embedding} \\
\href{https://huggingface.co/google/embeddinggemma-300m}{\texttt{google/embeddinggemma-300m}} & 2025-09 & 768 & \citet{Vera2025EmbeddingGemma} \\
\midrule
\multicolumn{4}{@{}l}{\textit{Image} (seen)}\\
\href{https://huggingface.co/timm/vgg16.tv_in1k}{\texttt{timm/vgg16.tv\_in1k}} & 2014-09 & 4096 & \citet{Simonyan2015VGG} \\
\href{https://huggingface.co/timm/inception_v3.tv_in1k}{\texttt{timm/inception\_v3.tv\_in1k}} & 2015-12 & 2048 & \citet{Szegedy2016InceptionV3} \\
\href{https://huggingface.co/timm/densenet121.tv_in1k}{\texttt{timm/densenet121.tv\_in1k}} & 2016-08 & 1024 & \citet{Huang2017DenseNet} \\
\href{https://huggingface.co/timm/resnext101_32x8d.tv_in1k}{\texttt{timm/resnext101\_32x8d.tv\_in1k}} & 2016-11 & 2048 & \citet{Xie2017ResNeXt} \\
\href{https://huggingface.co/timm/mobilenetv3_large_100.miil_in21k_ft_in1k}{\texttt{timm/mobilenetv3\_large\_100.miil\_in21k\_ft\_in1k}} & 2019-05 & 960 & \citet{Howard2019MobileNetV3} \\
\href{https://huggingface.co/google/vit-large-patch16-224-in21k}{\texttt{google/vit-large-patch16-224-in21k}} & 2020-10 & 1024 & \citet{Dosovitskiy2021AnII} \\
\href{https://huggingface.co/openai/clip-vit-large-patch14}{\texttt{openai/clip-vit-large-patch14}} & 2021-02 & 768 & \citet{radford2021learning} \\
\href{https://huggingface.co/timm/tf_efficientnetv2_l.in21k_ft_in1k}{\texttt{timm/tf\_efficientnetv2\_l.in21k\_ft\_in1k}} & 2021-04 & 1280 & \citet{Tan2021EfficientNetV2} \\
\href{https://huggingface.co/microsoft/swinv2-large-patch4-window12to16-192to256-22kto1k-ft}{\texttt{microsoft/swinv2-large-patch4-window12to16-192to256-22kto1k-ft}} & 2021-11 & 1536 & \citet{Liu2022SwinV2} \\
\href{https://huggingface.co/timm/vit_large_patch16_224.mae}{\texttt{timm/vit\_large\_patch16\_224.mae}} & 2021-11 & 1024 & \citet{He2022MAE} \\
\href{https://huggingface.co/facebook/convnext-large-224-22k-1k}{\texttt{facebook/convnext-large-224-22k-1k}} & 2022-01 & 1536 & \citet{Liu2022ConvNeXt} \\
\href{https://huggingface.co/timm/eva02_large_patch14_clip_224.merged2b}{\texttt{timm/eva02\_large\_patch14\_clip\_224.merged2b}} & 2023-03 & 1024 & \citet{Sun2023EVACLIP} \\
\href{https://huggingface.co/google/siglip2-so400m-patch14-384}{\texttt{google/siglip2-so400m-patch14-384}} & 2023-03$^\dagger$ & 1152 & \citet{Tschannen2025SigLIP2} \\
\multicolumn{4}{@{}l}{\hspace{1em}\textit{held-out}}\\
\href{https://huggingface.co/apple/aimv2-huge-patch14-224}{\texttt{apple/aimv2-huge-patch14-224}} & 2024-11 & 1536 & \citet{Fini2025AIMv2} \\
\href{https://huggingface.co/timm/vit_pe_core_large_patch14_336.fb}{\texttt{timm/vit\_pe\_core\_large\_patch14\_336.fb}} & 2025-04 & 1024 & \citet{Bolya2025PerceptionEncoder} \\
\href{https://huggingface.co/facebook/dinov3-vitl16-pretrain-lvd1689m}{\texttt{facebook/dinov3-vitl16-pretrain-lvd1689m}} & 2025-08 & 1024 & \citet{Simeoni2025DINOv3} \\
\href{https://huggingface.co/Qwen/Qwen3-VL-Embedding-8B}{\texttt{Qwen/Qwen3-VL-Embedding-8B}} & 2026-01 & 4096 & \citet{Li2026Qwen3VLEmbedding} \\
\href{https://huggingface.co/jinaai/jina-embeddings-v5-omni-small}{\texttt{jinaai/jina-embeddings-v5-omni-small}} & 2026-05 & 1024 & \citet{Honicke2026JinaV5Omni} \\
\midrule
\multicolumn{4}{@{}l}{\textit{Audio} (seen)}\\
\href{https://github.com/s3prl/s3prl}{\texttt{s3prl/vq\_wav2vec}}$^*$ & 2019-10 & 512 & \citet{Baevski2020VQWav2vec} \\
\href{https://huggingface.co/facebook/wav2vec2-large}{\texttt{facebook/wav2vec2-large}} & 2020-06 & 1024 & \citet{Baevski2020wav2vec2A} \\
\href{https://github.com/s3prl/s3prl}{\texttt{s3prl/tera}}$^*$ & 2020-07 & 768 & \citet{Liu2021TERA} \\
\href{https://huggingface.co/MIT/ast-finetuned-audioset-10-10-0.4593}{\texttt{MIT/ast-finetuned-audioset-10-10-0.4593}} & 2021-04 & 768 & \citet{Gong2021AST} \\
\href{https://huggingface.co/facebook/hubert-large-ll60k}{\texttt{facebook/hubert-large-ll60k}} & 2021-06 & 1024 & \citet{Hsu2021HuBERTSS} \\
\href{https://huggingface.co/Simon-Kotchou/ssast-base-patch-audioset-16-16}{\texttt{Simon-Kotchou/ssast-base-patch-audioset-16-16}} & 2021-10 & 768 & \citet{Gong2022SSAST} \\
\href{https://huggingface.co/microsoft/wavlm-large}{\texttt{microsoft/wavlm-large}} & 2021-10 & 1024 & \citet{Chen2022WavLM} \\
\href{https://huggingface.co/facebook/wav2vec2-xls-r-300m}{\texttt{facebook/wav2vec2-xls-r-300m}} & 2021-11 & 1024 & \citet{Babu2022XLSR} \\
\href{https://huggingface.co/facebook/data2vec-audio-large}{\texttt{facebook/data2vec-audio-large}} & 2022-02 & 1024 & \citet{Baevski2022Data2vec} \\
\href{https://huggingface.co/facebook/wav2vec2-conformer-rel-pos-large}{\texttt{facebook/wav2vec2-conformer-rel-pos-large}} & 2022-06 & 1024 & \citet{Wang2020FairseqS2T} \\
\href{https://huggingface.co/laion/larger_clap_general}{\texttt{laion/larger\_clap\_general}} & 2022-11 & 512 & \citet{Wu2023CLAP} \\
\href{https://github.com/microsoft/unilm/blob/master/beats/README.md}{\texttt{microsoft/unilm/BEATs\_iter3\_plus\_AS2M}}$^*$ & 2022-12 & 768 & \citet{Chen2023BEATs} \\
\href{https://huggingface.co/openai/whisper-large-v3}{\texttt{openai/whisper-large-v3}} & 2022-12 & 1280 & \citet{Radford2023RobustSR} \\
\href{https://huggingface.co/worstchan/EAT-base_epoch30_pretrain}{\texttt{worstchan/EAT-base\_epoch30\_pretrain}} & 2024-01 & 768 & \citet{Chen2024EAT} \\
\href{https://huggingface.co/mispeech/dasheng-0.6B}{\texttt{mispeech/dasheng-0.6B}} & 2024-06 & 1280 & \citet{Dinkel2024Dasheng} \\
\multicolumn{4}{@{}l}{\hspace{1em}\textit{held-out}}\\
\href{https://huggingface.co/tsinghua-ee/WAVE-7B}{\texttt{tsinghua-ee/WAVE-7B}} & 2025-09 & 3584 & \citet{Tang2025WAVE} \\
\href{https://huggingface.co/LCO-Embedding/LCO-Embedding-Omni-7B}{\texttt{LCO-Embedding/LCO-Embedding-Omni-7B}} & 2025-10 & 3584 & \citet{Xiao2025LCOEmbedding} \\
\href{https://huggingface.co/nvidia/omni-embed-nemotron-3b}{\texttt{nvidia/omni-embed-nemotron-3b}} & 2025-10 & 2048 & \citet{Xu2025OmniEmbedNemotron} \\
\href{https://huggingface.co/jinaai/jina-embeddings-v5-omni-small}{\texttt{jinaai/jina-embeddings-v5-omni-small}} & 2026-05 & 1024 & \citet{Honicke2026JinaV5Omni} \\
\href{https://huggingface.co/MIT-SLS/USAD2-Large}{\texttt{MIT-SLS/USAD2-Large}} & 2026-06 & 1024 & \citet{Chang2026USAD2} \\
\bottomrule
\end{tabular}
}
\end{table}

\section{\model{} Meta-Training Details}
\label{app:model-architecture}
\label{app:architecture}

Table~\ref{tab:repicl-config} lists the hyperparameters of \model{}; the whitening and architecture follow Section~\ref{sec:method}. Each fold is meta-trained once on a single NVIDIA RTX 4090; for the audio fold this takes about 31 hours for \model{}-I and about 45 hours for \model{}-T.

\begin{table}[!htbp]
\caption{Hyperparameters of \model{}-I and \model{}-T.}
\label{tab:repicl-config}
\centering
\small
\setlength{\tabcolsep}{6pt}
\begin{tabular}{@{}l>{\centering\arraybackslash}p{2.4cm}>{\centering\arraybackslash}p{2.4cm}@{}}
\toprule
 & \model{}-I & \model{}-T \\
\midrule
\multicolumn{3}{@{}l}{\textit{Episode and whitening}}\\
Way / shot / queries per class & \multicolumn{2}{c}{5 / 5 / 10} \\
Fit set $\mathcal{A}$ & $\mathcal{Z}_\mathcal{S}$ & $\mathcal{Z}_\mathcal{S} \cup \mathcal{Z}_\mathcal{Q}$ \\
PCA rank $r_{\mathcal{A}}$ & 24 & 74 \\
Standardization $\varepsilon$ & \multicolumn{2}{c}{$10^{-5}$} \\
\midrule
\multicolumn{3}{@{}l}{\textit{Architecture}}\\
Layers / heads & \multicolumn{2}{c}{24 / 16} \\
Token dimension (feature + label) & \multicolumn{2}{c}{1024 (768 + 256)} \\
MLP dimension & \multicolumn{2}{c}{4096} \\
Attention mask $\bm{M}$ & queries masked & none \\
Readout temperature $\tau$ & \multicolumn{2}{c}{0.07} \\
Parameters & \multicolumn{2}{c}{$\approx$303M} \\
Initialization & \multicolumn{2}{c}{CAML-CLIP} \\
\midrule
\multicolumn{3}{@{}l}{\textit{Meta-training}}\\
Optimizer & \multicolumn{2}{c}{AdamW} \\
Learning rate / warmup steps & \multicolumn{2}{c}{$10^{-4}$ / 2{,}500} \\
Weight decay & \multicolumn{2}{c}{0.01} \\
Optimizer steps & \multicolumn{2}{c}{50{,}000} \\
Episodes per step / microbatch & \multicolumn{2}{c}{1{,}024 / 128} \\
Input noise $\sigma$ & \multicolumn{2}{c}{0.07} \\
EMA decay & \multicolumn{2}{c}{0.999} \\
Precision & \multicolumn{2}{c}{bf16} \\
Seed & \multicolumn{2}{c}{42} \\
\bottomrule
\end{tabular}
\end{table}

\paragraph{Task sampling.}
Each fold meta-trains on the training tasks of its two non-held-out modalities. Each optimizer step draws 1{,}024 training tasks. The two modalities receive equal probability; within a modality, a dataset $\mathcal{D}$ is drawn with probability proportional to $\sqrt{\tilde n_\mathcal{D}}$, where $\tilde n_\mathcal{D}$ is its training-pool size clipped to $[500,\,200{,}000]$; and the encoders paired with $\mathcal{D}$ are equally likely. A task $(\mathcal{D},E)$ of modality $m$ is thus drawn with probability
\[
p(\mathcal{D},E)=\frac12\cdot
\frac{\sqrt{\tilde n_\mathcal{D}}}{\sum_{\mathcal{D}'\in\mathcal{D}_m}\sqrt{\tilde n_{\mathcal{D}'}}}\cdot
\frac{1}{|\mathcal{E}_D|},
\]
where $\mathcal{D}_m$ is the set of training datasets of $m$ and $\mathcal{E}_\mathcal{D}$ the set of encoders paired with $\mathcal{D}$. The square root favors larger datasets without letting them dominate. Each drawn task yields one episode, whose class labels are randomly permuted.

\paragraph{Optimization.}
We minimize $\mathcal{L}(\theta)$ of Section~\ref{subsec:training} with AdamW (learning rate $10^{-4}$, weight decay 0.01) for 50{,}000 steps; the learning rate warms up linearly over the first 2{,}500 steps and then stays constant. The 1{,}024 episodes of a step are processed as 8 accumulated microbatches of 128. As a simple input augmentation against overfitting to the training tasks, Gaussian noise with standard deviation $\sigma=0.07$ is added to the $r_{\mathcal{A}}$ active coordinates of each whitened token during training. Model weights are kept in bf16, while the whitening statistics are computed in fp32.

\paragraph{Weight averaging and checkpoint selection.}
To smooth the step-to-step noise of episodic updates, we keep an exponential moving average $\bar\theta$ of the weights \citep{Polyak1992Averaging,MoralesBrotons2024EMA} with decay 0.999. Over the first 1{,}000 steps $t$, the decay is lowered to $(1+t)/(10+t)$ so that the initialization does not dominate the average. We evaluate $\bar\theta$ on the validation tasks every 2{,}500 steps and keep the checkpoint with the highest validation accuracy, which is used for every reported result.

\section{Comparison Methods}
\label{app:comparison-details}

For all methods other than \model{}, we use the raw frozen representation without episodic whitening.

\subsection{Classical Classifiers Fitted on the Support Set}

Logistic regression and an RBF-kernel SVM serve as a linear and a nonlinear classifier fitted on the support set of each episode. Both use scikit-learn 1.6.1 \citep{Pedregosa2011Scikit} at its defaults after StandardScaler normalization fitted on the support set, except that logistic regression runs for up to 1{,}000 iterations.

\subsection{Few-Shot Methods on Frozen Representations}

\paragraph{ProtoHead.}
ProtoHead is the variant of prototypical networks \citep{Snell2017PrototypicalNF} on a frozen backbone used by \citet{Woerner2026UniversalAL}: each query is assigned to the class whose mean $L_2$-normalized support representation is nearest.

\paragraph{SimpleShot.}
SimpleShot \citep{Wang2019SimpleShot} applies nearest-centroid classification after a feature transformation. Its strongest variant, CL2N, centers with the mean of the encoder's base-class training features, which are unavailable for our encoders; it therefore centers with the support mean, $L_2$-normalizes, and scores by cosine similarity to the class means.

\paragraph{LaplacianShot and $\alpha$-TIM.}
LaplacianShot \citep{Ziko2020LaplacianShot} refines nearest-prototype predictions with a Laplacian term that encourages nearby queries to share a label, and $\alpha$-TIM \citep{Veilleux2021RealisticTransductive} maximizes the mutual information between queries and their predictions with an $\alpha$-divergence suited to class-imbalanced query sets. LaplacianShot uses easyfsl 1.5.0 \citep{Bennequin2021EasyFSL} at its defaults, and $\alpha$-TIM uses $\alpha=7$, temperature 15, and 1{,}000 gradient steps with learning rate $10^{-4}$. Both take $L_2$-normalized inputs.

\paragraph{PT-MAP.}
PT-MAP \citep{Hu2021PTMAP} applies a power transform to the features and then estimates the class centers jointly from the support and query sets with an optimal-transport-based MAP procedure. It follows the reference implementation, whose power transform $x^{1/2}$ assumes the nonnegative features of a post-ReLU backbone; since our representations are signed, it uses $\mathrm{sign}(x)\,|x|^{1/2}$.

\subsection{In-Context Learners}

\paragraph{TAIL.}
TAIL \citep{Woerner2026UniversalAL} is a transformer-based algorithm-implicit meta-learner that operates across tasks with varying domains, modalities, and label configurations. Its checkpoint is not released, so TAIL uses our reproduction, with inputs rescaled to norm 13 (Appendix~\ref{app:tail-reproduction}).

\paragraph{Tabular foundation models.}
Tabular foundation models are pretrained on large collections of synthetic tabular datasets and predict the labels of an unseen table in context, without gradient updates. TabPFN v2.6 \citep{Hollmann2025TabPFNv2} from tabpfn 7.0.1 and TabICL v2 \citep{Qu2026TabICLv2} from tabicl 2.0.2 take each support representation as a row of tabular features with its label as the target, and each query representation as a test row. TabPFN uses its \texttt{ignore\_pretraining\_limits} option to accept representations with more than 2{,}000 dimensions.

\FloatBarrier
\subsection{TAIL Reproduction and Adaptation}
\label{app:tail-reproduction}

\paragraph{Reproduction.}
TAIL \citep{Woerner2026UniversalAL} releases code but no checkpoint, so we train the model for 500 epochs on the CLIP ViT-H features of ImageNet-1k, Meta-Album, and MedIMeta, following their released recipe. Table~\ref{tab:tail-reproduction} compares our reproduction against the published numbers under TAIL's own task taxonomy; the two agree closely across all settings. One limitation: TAIL excludes a subset of classes from some datasets at evaluation time, but the released codebase does not implement this filtering, so our reproduction omits it and we cannot guarantee an identical evaluation protocol to theirs. As Table~\ref{tab:tail-reproduction} shows, however, the resulting gap remains small across all settings. All TAIL results reported in this paper, including on TAIL's own evaluation datasets, use this reproduced checkpoint.

\begin{table}[!htbp]
\caption{TAIL paper-reported and reproduced macro accuracy (\%).}
\label{tab:tail-reproduction}
\centering
\setlength{\tabcolsep}{4pt}
\small
\begin{tabular}{@{}lccccccc@{}}
\toprule
& \multicolumn{3}{c}{1-shot} & \multicolumn{3}{c}{5-shot} & \\
\cmidrule(lr){2-4}\cmidrule(lr){5-7}
Setting & Paper & Reprod. & Gap & Paper & Reprod. & Gap & Per-dataset MAE \\
\midrule
In-domain (4 datasets)     & 90.10 & 89.73 & -0.37 & 95.66 & 95.10 & -0.56 & 1.03 \\
Cross-modal (2 datasets)   & 62.75 & 64.31 & +1.56 & 72.48 & 72.53 & +0.05 & 1.00 \\
Cross-domain (8 datasets)  & 72.46 & 73.25 & +0.79 & 82.32 & 82.93 & +0.61 & 3.87 \\
\bottomrule
\end{tabular}
\end{table}

\paragraph{Transfer to \benchmark{}.}
When we evaluate TAIL directly on \benchmark{}, it collapses to near-random accuracy on most individual tasks (Table~\ref{tab:tail-fold}). Feature scale accounts for a substantial part of this loss. TAIL is trained on unnormalized CLIP features whose $L_2$ norm is about 13, whereas our representations are preprocessed to unit norm. Rescaling each embedding to a shared target norm recovers most of the loss, and a grid search on the DINOv3, USAD2, and E5-large-v2 tasks shows that performance peaks at a target norm of 13 for each encoder, close to the norm of TAIL's own training features (Table~\ref{tab:tail-scale}). The table also reports each encoder's original pre-normalization norm and TAIL accuracy on freshly re-encoded raw features; the latter preserves row-to-row magnitude variation and is distinct from a fixed target norm. We call this rescaling the scale calibration. Another limitation is TAIL's model design, which accepts at most the 1{,}280 dimensions it was trained on, so we pass wider representations through a fixed Gaussian projection, one untrained matrix per representation dimension. Table~\ref{tab:tail-width} separates each setting into the tasks that fit TAIL natively and those that pass through the projection. The gap between the two bands is large, but it is the same gap for \model{}-T, which takes every dimension natively, and the two agree in sign in all adaptation settings. The similar pattern for \model{}-T suggests that the projection is unlikely to be the primary explanation for TAIL's performance gap. We therefore apply the scale calibration and the projection to every TAIL result on \benchmark{}.

\begin{table}[!htbp]
\caption{TAIL variants and \model{}-T on \benchmark{}. TAIL (unit-norm):
raw unit-norm representations. TAIL (norm 13): scale-calibrated to norm 13.
TAIL\textsuperscript{\(\star\)}: retrained on our folds' training tasks.
\model{}-T is shown for reference. Bold marks the best TAIL variant per
column.}
\label{tab:tail-fold}
\centering
\footnotesize
\setlength{\tabcolsep}{3.5pt}
\begin{tabular}{@{}lcccccccccccc@{}}
\toprule
& \multicolumn{4}{c}{\textbf{Text-held}} & \multicolumn{4}{c}{\textbf{Audio-held}} & \multicolumn{4}{c}{\textbf{Image-held}} \\
\cmidrule(lr){2-5}\cmidrule(lr){6-9}\cmidrule(l){10-13}
\textbf{Method} & D & E & J & M & D & E & J & M & D & E & J & M \\
\midrule
TAIL (unit-norm)         & 26.86 & 34.48 & 29.01 & 28.54 & 30.34 & 34.67 & 32.42 & 24.27 & 25.82 & 29.15 & 26.14 & 38.22 \\
TAIL (norm 13)           & \textbf{52.79} & \textbf{71.96} & \textbf{61.05} & \textbf{62.23} & \textbf{66.06} & \textbf{71.13} & \textbf{71.07} & \textbf{45.67} & \textbf{50.33} & \textbf{64.93} & \textbf{55.55} & \textbf{79.31} \\
TAIL\textsuperscript{\(\star\)} & 36.96 & 48.48 & 38.86 & 33.07 & 55.43 & 62.33 & 60.21 & 36.75 & 21.81 & 20.16 & 20.08 & 20.25 \\
\midrule
\model{}-T               & 63.55 & 77.99 & 68.29 & 73.26 & 75.27 & 78.13 & 77.47 & 55.62 & 61.79 & 76.58 & 65.02 & 85.69 \\
\bottomrule
\end{tabular}
\end{table}

\begin{table*}[t]
\caption{TAIL accuracy (\%) across input scales. ``Raw'' uses unnormalized
features; numeric columns indicate target norms. ``Raw norm'' is the mean
pre-$L_2$ embedding norm.}
\label{tab:tail-scale}
\centering
\small
\setlength{\tabcolsep}{4pt}
\begin{tabular*}{\textwidth}{@{\extracolsep{\fill}}llcccccccccc@{}}
\toprule
& & & & \multicolumn{7}{c}{TAIL input scale} & \\
\cmidrule(lr){5-11}
Encoder & Modality & \# tasks & Raw norm & Raw & 1 & 3 & 6 & 13 & 26 & 52 & Ours \\
\midrule
DINOv3 & image & 19 & 15.61 & 87.50 & 50.65 & 83.50 & 87.00 & \textbf{87.62} & 84.75 & 63.15 & 92.34 \\
USAD2 & audio & 21 & 163.86 & 6.59 & 21.98 & 31.21 & 44.85 & \textbf{51.00} & 40.25 & 23.77 & 63.94 \\
E5-large-v2 & text & 19 & 23.03 & 65.48 & 24.40 & 49.02 & 66.94 & \textbf{72.44} & 61.37 & 29.09 & 82.82 \\
\bottomrule
\end{tabular*}
\end{table*}

\paragraph{Training TAIL on \benchmark{}.}
The scale calibration and projection correct the representation mismatch, but TAIL's training tasks still differ from \model{}'s. \model{} is meta-trained on the standard tasks of each fold, so for a fair comparison we also retrain TAIL using \benchmark{}'s meta-training setup and write the model as TAIL\textsuperscript{\(\star\)}. The results in Table~\ref{tab:tail-fold} show that matching the training setup does not help. TAIL\textsuperscript{\(\star\)} loses to both the scale-calibrated TAIL and \model{}-T. The main results therefore use the scale-calibrated TAIL. One limit bounds this reading. We transfer TAIL's published hyperparameters rather than searching over them, so a tuned TAIL\textsuperscript{\(\star\)} may score higher.

\begin{table}[!htbp]
\caption{Accuracy (\%) split by input width: native is at most 1{,}280
dimensions, where TAIL fits without modification; proj.\ covers dimensions
larger than 1{,}280, passed through the fixed Gaussian matrix.}
\label{tab:tail-width}
\centering
\footnotesize
\setlength{\tabcolsep}{2pt}
\begin{tabular}{@{}lcccccccccccc@{}}
\toprule
& \multicolumn{4}{c}{\textbf{Text-held}} & \multicolumn{4}{c}{\textbf{Audio-held}} & \multicolumn{4}{c}{\textbf{Image-held}} \\
\cmidrule(lr){2-5}\cmidrule(lr){6-9}\cmidrule(l){10-13}
\textbf{Width} & D & E & J & M & D & E & J & M & D & E & J & M \\
\midrule
\multicolumn{13}{@{}l}{\textbf{TAIL (norm 13)}} \\
\quad native   & 49.58 & 76.94 & 68.16 & 57.94 & 63.47 & 86.18 & 86.36 & 42.40 & 46.61 & 64.72 & 55.74 & 82.01 \\
\quad proj.    & 74.80 & 66.91 & 54.67 & 67.29 & 70.40 & 60.81 & 61.78 & 64.20 & 67.89 & 65.02 & 55.46 & 75.02 \\
\quad $\Delta$ & $+$25.22 & $-$10.03 & $-$13.49 & $+$9.35 & $+$6.93 & $-$25.37 & $-$24.58 & $+$21.80 & $+$21.28 & $+$0.30 & $-$0.28 & $-$6.99 \\
\midrule
\multicolumn{13}{@{}l}{\textbf{\model{}-T}} \\
\quad native   & 60.61 & 83.72 & 75.92 & 68.64 & 73.48 & 90.21 & 91.14 & 53.11 & 58.23 & 75.55 & 65.07 & 89.34 \\
\quad proj.    & 83.62 & 72.18 & 61.45 & 78.71 & 78.28 & 69.84 & 69.18 & 69.84 & 78.51 & 77.01 & 64.98 & 79.89 \\
\quad $\Delta$ & $+$23.01 & $-$11.54 & $-$14.47 & $+$10.07 & $+$4.80 & $-$20.37 & $-$21.96 & $+$16.73 & $+$20.28 & $+$1.46 & $-$0.09 & $-$9.45 \\
\bottomrule
\end{tabular}
\end{table}

\section{Detailed Results of the Main Paper}
\label{app:detailed-results}

\subsection{Per-Dataset Results under the TAIL Protocol}
\label{app:tail-benchmark}

Table~\ref{tab:tail-bench-full} gives the detailed results summarized in Section~\ref{subsec:main}. The evaluation follows TAIL's setup, with CLIP features and 1{,}000 identical 5-way 5-shot episodes per dataset. Both \model{} variants use the image-held-out checkpoint, so the 11 datasets and the CLIP image encoder are unseen during meta-training, whereas the CLIP text encoder is among the training encoders. CAML and GPICL are included through their published results, since their released formulations do not support the heterogeneous input dimensions of \benchmark{}. Both TAIL and our main experiments include a linear baseline, but the two fit it differently. The table therefore reports three Linear columns: the numbers published by TAIL, our reproduction of the TAIL recipe, and our logistic regression. TAIL fits an unregularized linear layer per episode with 100 epochs of Adam, while ours fits a standardized, L2-regularized logistic regression. TAIL's recipe is therefore not necessarily the strongest possible per-episode linear classifier. \model{}-T outperforms every previously reported result on average, reaching 92.31\% against TAIL's own best reported number of 90.05\%.

\begin{table}[!htbp]
\caption{Accuracy (\%) on the visual datasets of TAIL. $^\dagger$ and the
published columns are values reported by \citet{Woerner2026UniversalAL}. Bold
marks the best result per row, excluding OCT (4-way).}
\label{tab:tail-bench-full}
\centering
\scriptsize
\setlength{\tabcolsep}{3.5pt}
\begin{tabular}{@{}lccccccccc@{}}
\toprule
& \multicolumn{3}{c}{Linear} & & & \multicolumn{2}{c}{TAIL} & \multicolumn{2}{c}{Ours} \\
\cmidrule(lr){2-4}\cmidrule(lr){7-8}\cmidrule(l){9-10}
Dataset & published & TAIL recipe & ours & GPICL$^\dagger$ & CAML$^\dagger$ & published & reproduced & \model{}-I & \model{}-T \\
\midrule
CIFAR-FS         & 91.86 & 91.56 & 92.87 & 91.20 & 91.69 & 94.55 & 94.49 & 93.75 & \textbf{96.75} \\
miniImageNet     & 97.65 & 97.66 & 98.56 & 99.44 & 99.29 & 99.63 & 99.40 & 98.71 & \textbf{99.77} \\
tieredImageNet   & 95.30 & 95.12 & 96.50 & 98.18 & 97.98 & \textbf{98.67} & 98.05 & 96.72 & 98.43 \\
Pascal VOC       & 83.57 & 83.49 & 85.70 & 87.46 & 87.87 & \textbf{89.78} & 88.33 & 86.13 & 89.55 \\
Aircraft         & 92.12 & 91.82 & 93.65 & 90.83 & 93.09 & 95.01 & 94.31 & 94.12 & \textbf{96.66} \\
CUB              & 95.17 & 94.77 & 96.81 & 96.02 & 97.55 & 98.51 & 97.52 & 97.41 & \textbf{98.93} \\
meta-iNat        & 92.91 & 92.91 & 94.78 & 95.56 & 96.20 & \textbf{97.70} & 96.77 & 95.37 & 97.66 \\
tiered meta-iNat & 88.32 & 88.80 & 92.38 & 91.15 & 93.53 & 95.69 & 94.24 & 93.41 & \textbf{96.56} \\
Paintings        & 66.21 & 66.69 & 68.85 & 54.29 & 58.72 & 63.03 & 73.39 & 69.93 & \textbf{73.82} \\
Pascal-Paintings & 71.62 & 71.10 & 74.78 & 65.79 & 70.13 & 73.04 & 79.82 & 76.34 & \textbf{81.81} \\
PBC              & 68.39 & 68.01 & 76.67 & 53.28 & 80.19 & 84.96 & 78.90 & 79.13 & \textbf{85.41} \\
Mean             & 85.74 & 85.63 & 88.32 & 83.93 & 87.84 & 90.05 & 90.47 & 89.18 & \textbf{92.31} \\
OCT (4-way)      & --    & 49.82 & 54.64 & -- & -- & --    & 47.94 & 54.07 & 55.94 \\
\bottomrule
\end{tabular}
\end{table}

\subsection{Full Ablation Results}
\label{app:ablation-full}

Table~\ref{tab:ablation-full} gives the detailed results of the ablation study in Section~\ref{subsec:ablation} for every fold and adaptation setting. For meta-training diversity, we ablate the set of meta-training tasks instead of a model component, to see how performance depends on the diversity of that set. Each fold meta-trains on two modalities, and we restrict the task pool $\mathcal{T}_{\mathrm{train}}$ within each of them, keeping either one dataset with all its compatible encoders or one encoder with all its compatible datasets. Both settings are trained for the same 50{,}000 steps as the full model.

With limited task diversity, the choice of dataset or encoder matters. If the selected dataset is one on which few-shot classification is hard, its episodes may provide only a weak training signal, and the resulting drop would confound the effect of diversity with the difficulty of that dataset. The same applies to the selected encoder. We therefore select the source favorably. In each training modality, we choose, among the training datasets, the dataset with the highest Logistic Regression accuracy averaged over its compatible encoders, and, among the training encoders, the encoder with the highest accuracy averaged over its compatible datasets. For text, image, and audio, the selected datasets are HWU64, Oxford Flowers, and ESC-50, and the selected encoders are BGE-ICL, EVA02-CLIP-L, and Whisper-large-v3.

This selection has two limitations. First, it uses Logistic Regression accuracy on the test split, which may favor the restricted settings. Second, because the restriction is applied within each modality, every restricted setting still contains two datasets or two encoders. The settings therefore reduce the diversity of the meta-training pool rather than remove it, and we do not test more restricted pools, such as a single dataset or encoder in total.

\begin{table}[!htbp]
\caption{Detailed results of the ablation study.}
\label{tab:ablation-full}
\centering
\scriptsize
\setlength{\tabcolsep}{2.5pt}
\renewcommand{\arraystretch}{1.05}
\begin{tabular}{@{}lcccccccccccc@{}}
\toprule
& \multicolumn{4}{c}{\textbf{Text-held-out fold}}
& \multicolumn{4}{c}{\textbf{Audio-held-out fold}}
& \multicolumn{4}{c}{\textbf{Image-held-out fold}} \\
\cmidrule(lr){2-5}\cmidrule(lr){6-9}\cmidrule(l){10-13}
& D & E & J & M & D & E & J & M & D & E & J & M \\
\midrule
\textbf{\model{}-I}
& 61.41 & 75.85 & 66.22 & 70.10 & 73.52 & 76.79 & 75.86 & 55.14 & 59.19 & 72.82 & 62.56 & 83.26 \\
\quad\emph{Episodic whitening} \\
\qquad w/o Coord. Standardize
& $-$8.18 & $-$3.64 & $-$5.19 & $-$7.88
& $-$5.03 & $-$4.14 & $-$3.40 & $-$9.38
& $-$7.98 & $-$6.53 & $-$5.85 & $-$3.25 \\
\qquad w/o PCA, trainable proj.
& $-$5.28 & $-$5.94 & $-$7.15 & $-$12.07
& $-$8.61 & $-$11.38 & $-$10.76 & $-$9.93
& $-$5.25 & $-$12.39 & $-$10.34 & $-$6.55 \\
\qquad w/o PCA, fixed proj.
& $-$5.44 & $-$6.05 & $-$8.37 & $-$10.96
& $-$8.33 & $-$10.60 & $-$10.06 & $-$10.01
& $-$5.33 & $-$11.88 & $-$9.96 & $-$6.42 \\
\quad\emph{Model structure} \\
\qquad w/o CAML initialization
& $-$0.08 & $-$0.01 & \textbf{+0.25} & $-$0.05
& $-$0.16 & $-$0.25 & $-$0.13 & $-$0.04
& \textbf{+0.02} & $-$0.10 & \textbf{+0.22} & \textbf{+0.02} \\
\qquad fixed five-class head
& $-$0.12 & $-$0.07 & \textbf{+0.38} & $-$0.14
& $-$0.28 & $-$0.45 & $-$0.14 & $-$0.04
& $-$0.22 & $-$0.45 & \textbf{+0.14} & $-$0.05 \\
\quad\emph{Meta-training diversity} \\
\qquad one dataset per modality
& $-$0.37 & $-$0.34 & \textbf{+0.03} & $-$0.74
& $-$0.40 & $-$0.32 & $-$0.14 & $-$0.32
& $-$0.54 & $-$0.75 & $-$0.26 & $-$0.21 \\
\qquad one encoder per modality
& $-$0.79 & $-$0.27 & $-$0.03 & $-$0.41
& $-$0.99 & $-$1.03 & $-$0.54 & $-$0.76
& $-$0.83 & $-$0.80 & $-$0.33 & $-$0.29 \\
\midrule
\textbf{\model{}-T}
& 63.55 & 77.99 & 68.29 & 73.26 & 75.27 & 78.13 & 77.47 & 55.62 & 61.79 & 76.58 & 65.02 & 85.69 \\
\quad\emph{Episodic whitening} \\
\qquad w/o Coord. Standardize
& $-$13.64 & $-$6.82 & $-$9.37 & $-$9.81
& $-$6.60 & $-$6.04 & $-$6.05 & $-$14.34
& $-$14.00 & $-$15.24 & $-$13.93 & $-$6.58 \\
\qquad w/o PCA, trainable proj.
& $-$7.30 & $-$8.02 & $-$9.45 & $-$15.15
& $-$8.35 & $-$11.21 & $-$10.51 & $-$9.91
& $-$7.01 & $-$15.61 & $-$12.45 & $-$8.59 \\
\qquad w/o PCA, fixed proj.
& $-$6.89 & $-$7.81 & $-$9.16 & $-$14.84
& $-$9.40 & $-$12.03 & $-$11.14 & $-$10.10
& $-$7.77 & $-$16.09 & $-$12.78 & $-$8.78 \\
\quad\emph{Model structure} \\
\qquad w/o CAML initialization
& +0.00 & \textbf{+0.13} & $-$0.01 & \textbf{+0.04}
& \textbf{+1.30} & \textbf{+1.04} & \textbf{+0.85} & \textbf{+1.53}
& $-$0.28 & $-$0.11 & $-$0.20 & $-$0.17 \\
\qquad fixed five-class head
& \textbf{+0.83} & \textbf{+0.53} & \textbf{+0.75} & \textbf{+0.76}
& \textbf{+1.89} & \textbf{+1.43} & \textbf{+1.34} & \textbf{+2.44}
& \textbf{+0.32} & \textbf{+0.78} & \textbf{+0.60} & \textbf{+0.35} \\
\quad\emph{Meta-training diversity} \\
\qquad one dataset per modality
& $-$2.02 & $-$1.51 & $-$1.67 & $-$2.55
& $-$0.87 & $-$0.88 & $-$0.75 & $-$1.02
& $-$2.85 & $-$3.52 & $-$2.86 & $-$1.50 \\
\qquad one encoder per modality
& $-$2.43 & $-$1.04 & $-$1.22 & $-$1.39
& $-$2.09 & $-$2.53 & $-$2.02 & $-$2.01
& $-$2.67 & $-$3.47 & $-$2.77 & $-$1.16 \\
\bottomrule
\end{tabular}
\end{table}

\section{Additional Analyses and Experiments}
\label{app:additional}

\subsection{Linear vs. Previous Few-Shot Methods}
\label{app:linear-vs-fewshot}

In our Table~\ref{tab:main} results on \benchmark{}, logistic regression outperforms almost every few-shot method, with PT-MAP the only method ahead of it, and only on the image-modality shift. Classical few-shot methods such as $\alpha$-TIM and PT-MAP were developed and tuned on CNN backbones (ResNet, WideResNet), and usually outperform linear methods there.
We therefore ask whether the gap we observe here reflects a mismatch between the encoders these methods were designed for and those evaluated on \benchmark{}, rather than evidence that the earlier methods are weaker than logistic regression.

\paragraph{Matched CNN backbones.}
On the exact ResNet-18, ResNet-34, and WRN-28-10 backbones used in the few-shot literature, evaluated under our benchmark's 5-way 5-shot image task, LaplacianShot, $\alpha$-TIM, and PT-MAP are all ahead of logistic regression on the same frozen features (Table~\ref{tab:resnet-diagnostic}). SimpleShot is the exception and is roughly on par, because our setting gives it no base-class-supervised training to draw its centering statistics from. On these matched backbones, previous few-shot methods do outperform logistic regression.

\begin{table}[h]
\caption{Previous few-shot methods vs.\ logistic regression on CNN
backbones. Accuracy (\%).}
\label{tab:resnet-diagnostic}
\centering
\small
\setlength{\tabcolsep}{4pt}
\begin{tabular}{@{}lccccc@{}}
\toprule
Backbone & Logistic regression & SimpleShot & LaplacianShot & $\alpha$-TIM & PT-MAP (signed-power) \\
\midrule
ResNet-18 & 77.17 & 77.05 & 77.88 & 77.83 & \textbf{80.47} \\
ResNet-34 & 76.75 & 76.71 & 77.20 & 77.39 & \textbf{79.11} \\
WRN-28-10 & 58.68 & 59.20 & 60.30 & 62.12 & \textbf{65.18} \\
\bottomrule
\end{tabular}
\end{table}

\paragraph{The benchmark's own encoder set.}
We then run the same comparison over every encoder of \benchmark{}, in all three modalities (Table~\ref{tab:encoder-class}). Across the benchmark's encoder classes, the advantage is heterogeneous: previous few-shot methods lead on several classes, whereas Logistic Regression leads on many others and on text and audio overall. Thus, no previous few-shot method consistently dominates across the full range of encoder families and modalities. These methods were therefore not designed with transfer across encoder families in mind, and none of them holds up across the full set. That is the gap \model{} is trained to close.

\begin{table}[h]
\caption{Logistic regression vs.\ previous few-shot methods on \benchmark{},
grouped by encoder representation class. Bold marks the best method in each
row.}
\label{tab:encoder-class}
\centering
\scriptsize
\setlength{\tabcolsep}{4pt}
\begin{tabular}{@{}llcccccc@{}}
\toprule
Modality & Encoder class & Num. of Enc. & Logistic regression & SimpleShot & LaplacianShot & $\alpha$-TIM & PT-MAP \\
\midrule
Image & CNN-like                 &  7 & 80.44 & 80.96 & 81.42 & 81.43 & \textbf{83.23} \\
      & ViT / vision-SSL         &  6 & \textbf{76.82} & 73.18 & 72.67 & 75.45 & 74.93 \\
      & CLIP-like contrastive    &  3 & 92.50 & 91.33 & 91.37 & 92.20 & \textbf{93.09} \\
      & Multimodal embedding     &  2 & 90.65 & 90.22 & 90.31 & 90.50 & \textbf{91.35} \\
      & \emph{All}               & 18 & 82.38 & 81.12 & 81.15 & 82.24 & \textbf{83.01} \\
      \midrule
\addlinespace[2pt]
Text  & Static word embeddings   &  2 & \textbf{61.35} & 55.28 & 55.30 & 59.41 & 58.92 \\
      & Pretrained LM hidden states & 4 & \textbf{50.62} & 38.50 & 35.45 & 45.86 & 44.07 \\
      & Purpose-built text embeddings & 7 & 79.74 & 80.01 & 81.54 & 78.19 & \textbf{83.34} \\
      & LLM/instruction embeddings & 4 & 79.95 & 80.51 & 81.48 & 76.98 & \textbf{83.15} \\
      & LLM hidden-state features &  5 & \textbf{66.92} & 57.86 & 55.36 & 62.00 & 61.42 \\
      & Vision-language contrastive & 2 & \textbf{66.56} & 60.79 & 60.17 & 62.92 & 65.18 \\
      & \emph{All}               & 24 & \textbf{69.62} & 64.90 & 64.42 & 66.39 & 68.65 \\
      \midrule
\addlinespace[2pt]
Audio & Speech-pretrained / ASR  &  9 & \textbf{49.53} & 42.92 & 40.74 & 47.33 & 42.81 \\
      & General-audio SSL / event models & 6 & \textbf{53.39} & 47.00 & 45.39 & 51.52 & 47.57 \\
      & Audio-text contrastive   &  1 & 56.41 & 55.76 & 56.01 & 56.13 & \textbf{56.65} \\
      & Audio/multimodal LLMs    &  4 & \textbf{66.67} & 64.17 & 63.91 & 65.31 & 65.44 \\
      & \emph{All}               & 20 & \textbf{54.46} & 49.04 & 47.53 & 52.62 & 49.46 \\
\bottomrule
\end{tabular}
\end{table}

\FloatBarrier
\subsection{LoRA Comparison}
\label{app:lora-comparison}

RepICL adapts to a task in context, without updating any weights. An alternative is to update the encoder on the support set at test time, a paradigm studied for few-shot classification with adapters \citep{Li2022TSA}, full-backbone fine-tuning \citep{Hu2022PushingTL}, cosine classifiers \citep{Chen2019CloserLook}, and broader PEFT sweeps \citep{Basu2024StrongBaselines}. We compare against LoRA \citep{Hu2022LoRA}, one of the most widely used standard PEFT techniques, which is also a strong few-shot baseline \citep{Basu2024StrongBaselines,Zanella2024CLIPLoRA}.

\paragraph{Setup.}
For each episode we re-initialize LoRA with rank 8, alpha 16 on the q/v attention projections and fit it on the support set with a prototype mean-cosine loss, similar to the LoRA + ProtoNCC recipe of \citet{Basu2024StrongBaselines}. For whisper-large and SigLIP-2 we tune only the encoder used to extract features. All tasks use AdamW with learning rate $3\times10^{-4}$ and 50 steps, selected on held-out episodes of a few datasets per modality with one encoder per modality. We compare RepICL-I and RepICL-T with LoRA, with logistic regression on the frozen features, and with a frozen baseline that classifies queries by cosine similarity to class prototypes of the unadapted encoder output. Due to computation limits, we evaluate 9 encoders (3 per modality) from \benchmark{}, chosen to cover different encoder sizes, on 30 sealed episodes per task disjoint from the selection episodes. Average time is measured on a single RTX 4090 over two 5-way 5-shot episodes, on 6 of the 9 encoders.

\begin{table}[h]
\caption{Performance compared to LoRA.}
\label{tab:lora-scale}
\centering
\scriptsize
\setlength{\tabcolsep}{4pt}
\begin{tabular}{@{}llccccccc@{}}
\toprule
Encoder & Modality & Params & Tasks & Frozen & LoRA & RepICL-I & RepICL-T & Logistic regression \\
\midrule
AST audioset-base & audio & 86.2M & 20 & 46.6 & 55.7 & 55.8 & \textbf{56.5} & 54.8 \\
HuBERT-large & audio & 315.4M & 20 & 46.9 & 51.1 & 57.2 & \textbf{57.7} & 56.3 \\
whisper-large & audio & 637.0M & 20 & 40.6 & 61.5 & 64.4 & \textbf{64.8} & 63.5 \\
all-mpnet-base-v2 & text & 109.5M & 19 & 75.2 & 79.5 & 77.17 & \textbf{80.72} & 77.16 \\
embedding-gemma-300m & text & 302.9M & 19 & 78.2 & 82.3 & 80.95 & \textbf{84.25} & 80.99 \\
stella-1.5B-v5 & text & 1.54B & 19 & 80.1 & \textbf{85.4} & 81.17 & 84.01 & 80.19 \\
DINOv3 ViT-L/16 & image & 303.1M & 19 & 85.9 & 91.5 & 90.29 & \textbf{92.34} & 88.33 \\
SigLIP-2 so400m/384 & image & 428.2M & 19 & 88.7 & 90.9 & 93.08 & \textbf{94.73} & 91.13 \\
PE-core-L & image & 317.2M & 19 & 88.2 & 92.9 & 92.94 & \textbf{94.85} & 90.27 \\
\midrule
\multicolumn{4}{l}{Average over encoders} & 70.0 & 76.8 & 77.0 & \textbf{78.9} & 75.9 \\
\multicolumn{4}{l}{Avg.\ time (ms/query)} & 7.67 & 418.90 & 7.81 & 7.92 & 7.75 \\
\bottomrule
\end{tabular}
\end{table}

\paragraph{Results.}
LoRA consistently improves over the frozen encoder on all nine encoders (Table~\ref{tab:lora-scale}), confirming the benefit of test-time adaptation. \model{}-I performs on par with LoRA, and \model{}-T outperforms it on almost every encoder. Since \model{} requires no test-time parameter update, it is also 33--68$\times$ faster per query. Overall, \model{}-I matches LoRA at a fraction of the test-time cost, and \model{}-T further outperforms it.

\subsection{Inference Cost}
\label{app:inference-timing}

We compare the per-episode inference time of \model{} with the other methods to assess whether it is efficient enough for practical use. For each method, we time one 5-way 5-shot episode with 25 support and 50 query examples on frozen representations, from normalization to prediction. \model{}, TAIL, TabPFN, and TabICL run on one NVIDIA RTX 4090, and the remaining methods run on the CPU. All methods are timed under the same protocol and none is specifically optimized for speed.

\begin{figure}[!htbp]
  \centering
  \includegraphics[width=0.75\textwidth]{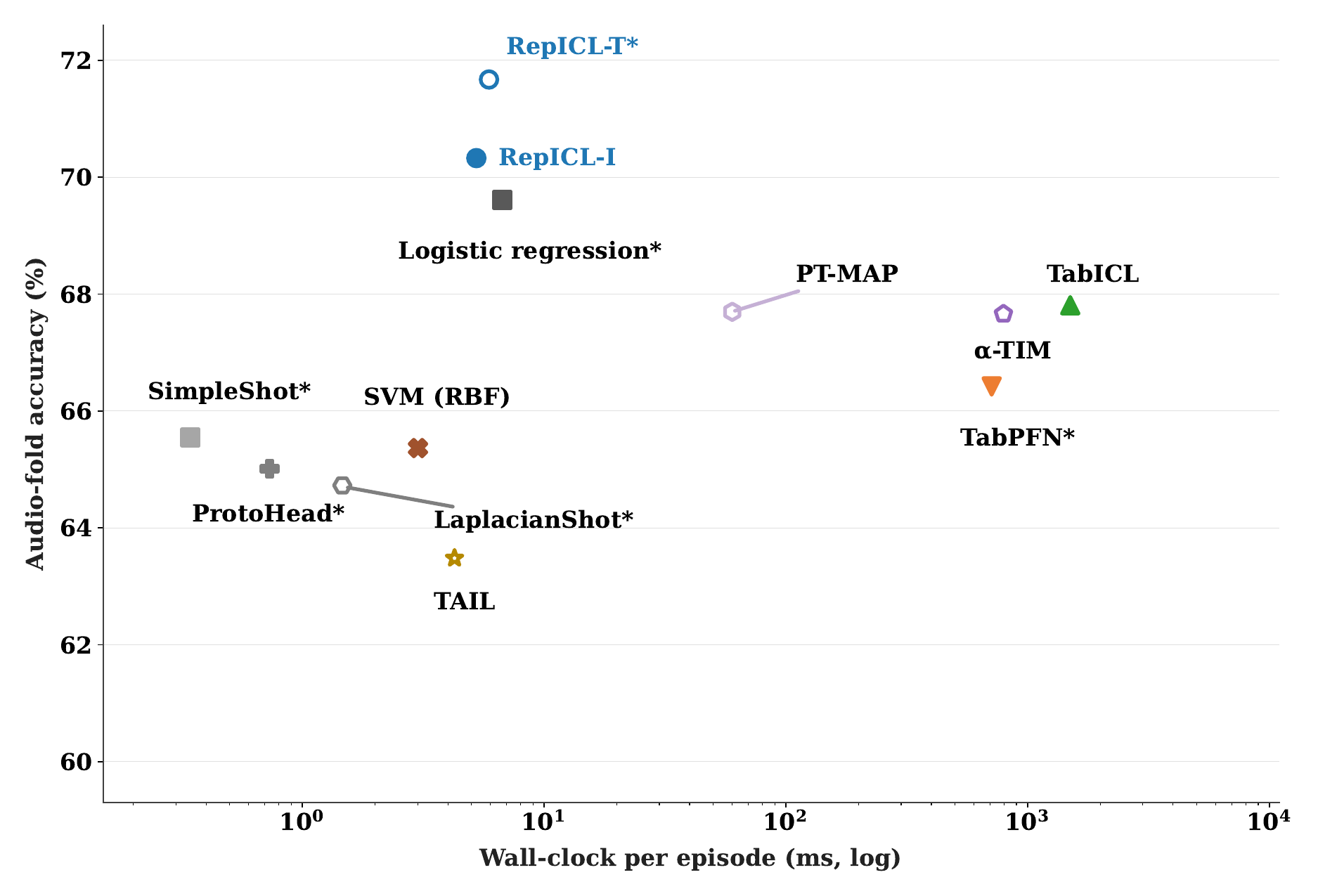}
  \caption{Accuracy against inference time per episode. Accuracy is the
  audio-fold mean from Table~\ref{tab:main}; hollow markers denote
  transductive methods. Time is the median over 22 episodes, one for each distinct
  representation width in the three folds. Methods marked with $^{*}$ show a
  run-to-run timing variation above 5\% on some episodes.}
  \label{fig:inference-timing}
\end{figure}

Figure~\ref{fig:inference-timing} shows that \model{} runs in a time comparable to TAIL and to Logistic Regression on the CPU, and is much faster than TabPFN and TabICL. Its median time per episode is 5.3~ms for \model{}-I and 5.9~ms for \model{}-T, against 6.7~ms for Logistic Regression, 0.7~s for TabPFN, and 1.5~s for TabICL. CPU and GPU times are not directly comparable, so these numbers indicate the order of magnitude of each method's inference time rather than a ranking.

\begin{figure}[!htbp]
  \centering
  \includegraphics[width=0.75\textwidth]{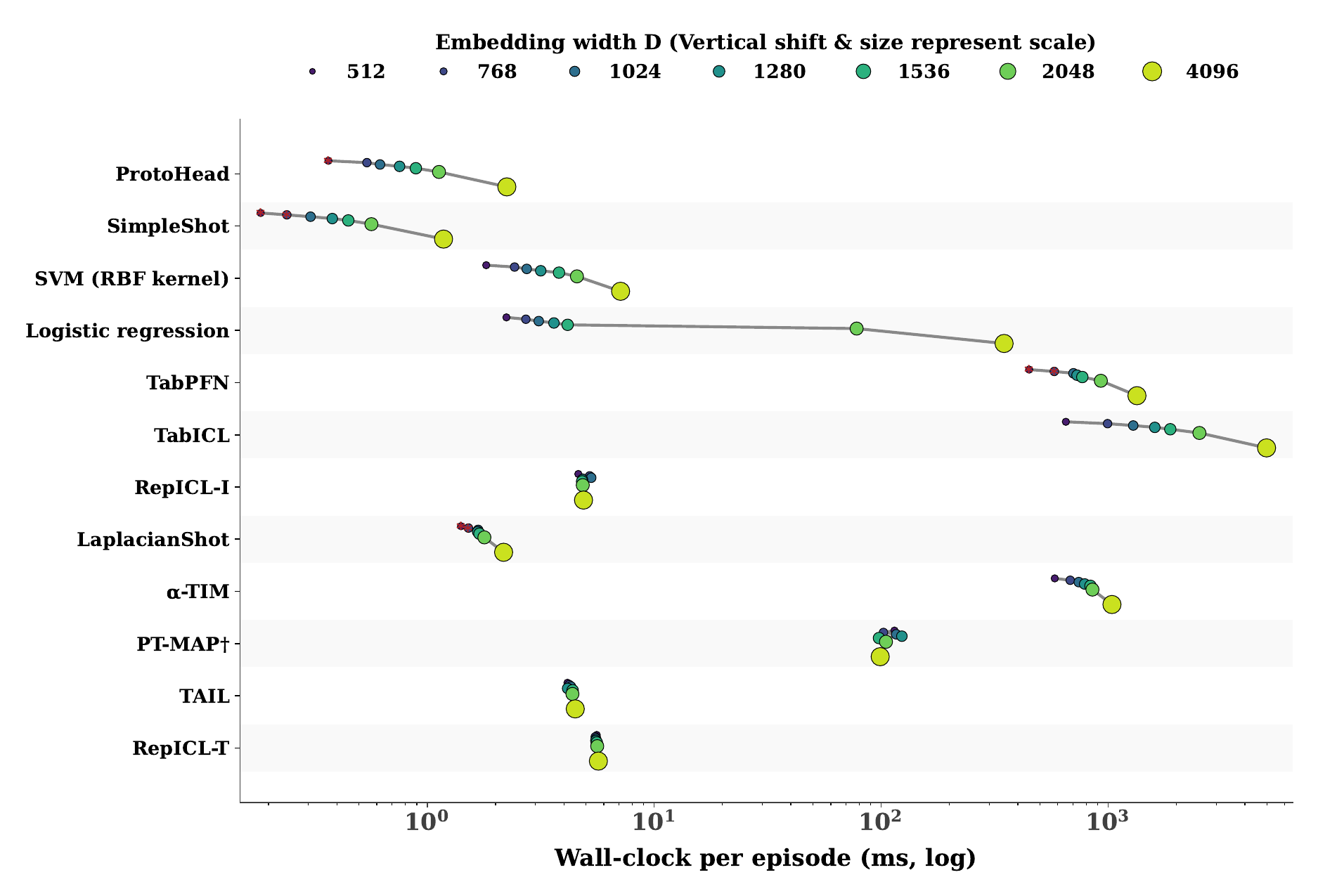}
  \caption{Inference time per episode as the representation width $D$ grows
  from 512 to 4096. PT-MAP$^{\dagger}$ is the median over 128 episodes, since its number of Sinkhorn iterations depends on the episode.}
  \label{fig:inference-width}
\end{figure}

The encoders in \benchmark{} produce representations of different widths, so we also examine how inference time depends on the representation width $D$. As Figure~\ref{fig:inference-width} shows, the inference time of \model{} is nearly independent of $D$, staying between 4.6 and 5.7~ms as $D$ grows from 512 to 4096, because only the episodic whitening operates in the $D$-dimensional space and the transformer always receives the same number of tokens.

\end{document}